\documentclass{article}
\usepackage{ijcai21}

\usepackage{times}
\usepackage{xcolor}
\usepackage{soul}
\usepackage{url}
\usepackage[hidelinks]{hyperref}
\usepackage[utf8]{inputenc}
\usepackage[small]{caption}
\usepackage{graphicx}
\usepackage{amsmath}
\usepackage{amsthm}
\usepackage{booktabs}
\usepackage{algorithm}
\usepackage{algorithmic}
\usepackage{hyperref}
\usepackage{graphicx}
\usepackage{caption}
\usepackage{float}
\usepackage{subcaption}
\usepackage{amsmath}
\usepackage{amsthm}
\usepackage{amssymb}
\usepackage{algorithm}
\usepackage{algorithmic}
\usepackage{hyperref}
\usepackage{breakcites}
\usepackage{etoolbox}

\newcommand{\citet}{\cite}
\newcommand{\citep}{\cite}
\providetoggle{proofs}
\settoggle{proofs}{false}

\usepackage{latexsym}

\newtheorem{theorem}{Theorem}

\newtheorem{lemma}[theorem]{Lemma}
\newtheorem{assumption}{Assumption}

\DeclareMathOperator*{\argmax}{arg\,max}

\newcommand{\E}{\mathbb{E}}
\newcommand{\KL}{D}

\title{Thompson Sampling for Bandits with Clustered Arms}

\author{
Emil Carlsson\footnote{Contact Author}\and
Devdatt Dubhashi \and
Fredrik D. Johansson
\affiliations
Department of Computer Science and Engineering, Chalmers University\\
\emails
\{caremil, dubhashi, fredrik.johansson\}@chalmers.se
}

\begin{document}
\maketitle

\begin{abstract}
We propose algorithms based on a multi-level Thompson sampling scheme, for the stochastic multi-armed bandit and its contextual variant with linear expected rewards, in the setting where arms are clustered. We show, both theoretically and empirically, how exploiting a given cluster structure can significantly improve the regret and computational cost compared to using standard Thompson sampling. In the case of the stochastic multi-armed bandit we give upper bounds on the expected cumulative regret showing how it depends on the quality of the clustering.  Finally, we perform an empirical evaluation showing that our algorithms perform well compared to previously proposed algorithms for bandits with clustered arms.  
\end{abstract}

\section{Introduction}
In a bandit problem, a learner must iteratively choose from a set of $N$  actions, also known as arms, in a sequence of $T$ steps as to minimize the expected cumulative regret over the horizon $T$ \citep{Lai1985}. Inherent in this setup is an exploration-exploitation tradeoff where the learner has to balance between exploring actions she is uncertain about in order to gain more information and exploiting current knowledge to pick actions that appears to be optimal.

In this work, we consider versions of the standard multi-armed bandit problem (MAB) and the contextual bandit with linear rewards (CB) where there is a clustering of the arms known to the learner. In the standard versions of these problems the cumulative regret scales with number of arms, $N$, which becomes problematic when the number of arms grows large \citep{Bubeck12}. Given a clustering structure one would like the exploit it to remove the explicit dependence on $N$ and replace it with a dependence on the given clustering instead. A motivating example is recommender systems in  e-commerce where there may be a vast amount of products organized into a much smaller set of categorizes. Users my have strong preferences for certain categorizes which yields similar expected rewards for recommending products from the same category. 

\paragraph{Our Contributions.} We propose algorithms based on a multi-level Thompson sampling \citep{Thompson1933} scheme for the stochastic multi-armed bandit with clustered arms (MABC) and its contextual variant with linear expected rewards and clustered arms (CBC). For the MABC, we provide regret bounds for our algorithms which completely removes the explicit dependence on $N$ in favor for a dependence on properties of the given clustering. We perform an extensive empirical evaluation showing both how the quality of the clustering affects the regret and that our algorithms are very competitive with recent algorithms proposed for MABC and CBC. Noteworthy is that the empirical evaluation shows that our algorithms still performs well even in the case where our theoretical assumptions are violated. 

\section{Stochastic Multi-armed Bandit with Clustered Arms}\label{sec:MABC}
We consider the MABC. As in the standard MAB problem we have a set of arms $\mathcal{A}$ of cardinality $N$. At each time step $t>0$ the learner must pick an arm $a_t \in \mathcal{A}$ after which an instant stochastic reward, $r_t(a_t)$, drawn from some distribution, $r_t \sim \mathcal{D}_{a_t}$, with an unknown mean $\E_{\mathcal{D}_{a_t}}[r_t] = \mu_{a_t}$. The goal of the learner is to maximize its expected cumulative reward  over a sequence of $T$ time steps or equivalently, to minimize its expected cumulative regret $\E[R_T]$ w.r.t the optimal arm $a^* = \argmax_{a \in \mathcal{A}} \mu_a$ in hindsight, $ R_T := \sum_{t=1}^T r_t(a^*) - r_t(a_t)$. 

In the MABC, the learner has, in addition, access to a clustering of the $N$ arms which may be used to guide exploration. We will consider two types of clustering: \begin{description}
    \item[Disjoint Clusters]  The $N$ arms are partitioned into a a set of clusters $\mathcal{K}$ such that each arm $a \in \mathcal{A}$ is associated to exactly one cluster.
    \item[Hierarchical Clustering] The $N$ arms are organized into a tree $\mathcal{T}$ of depth $L$ such that each arm is associated with a unique leaf of the tree.
\end{description}

We will show in Section \ref{sec:regret_TSC} and \ref{sec:regret_HTS} that when rewards are drawn from Bernoulli distributions, $r_t \sim \mathcal{B}(\mu_a)$, with unknown parameters $\mu_a$, the learner can exploit the known clustering to greatly improve the expected cumulative regret compared to the regret achievable with no knowledge of the cluster structure (under certain assumptions on the quality of the clustering).

\subsection{Thompson Sampling for MABC}\label{sec:TS_MABC}
In the celebrated Thompson sampling (TS) algorithm for MAB with Bernoulli distributed rewards a learner starts at time $t=0$ with a prior belief  $\text{Beta}(1, 1)$  over possible expected rewards, $\theta_a \in [0, 1]$,  for each $a \in A$.  At time $t$,  having observed  $S_t(a)$  number of successful $(r=1)$ plays and $F_t(a)$ the number of unsuccessful $(r=0)$ plays of arm $a$,  the learner's posterior belief over possible expected rewards for arm $a$ is $\text{Beta}(S_t(a), F_t(a))$,  where $S_0(a)=F_0(a)=1$.  At each time step $t$,  the learner samples an expected reward for each arm $\theta_a \sim \text{Beta}(S_t(a), F_t(a))$ and then acts greedily w.r.t. the sample means, i.e. the learner plays  the arm $a_t = \argmax_{a \in \mathcal{A}} \theta_a$.  Given a reward $r_t$ the learner updates the posterior of the played arm $a_t$ as $S_{t+1}(a_t) = S_{t} + r_t$ and $F_{t+1}(a_t) = F_t(a_t) + (1-r_t)$,. The posteriors of the arms not played are not updated.
\begin{algorithm}[t]
\caption{TSC}\label{alg:TSC}
\begin{algorithmic}
\REQUIRE $\mathcal{A}$, $\mathcal{K}$
\STATE Set $S_0 = F_0 = 1$ for all $a$ and $C$.
\FOR{$t=1, ..., T$}
    \STATE For each cluster $C$ sample $\theta_C \sim \text{Beta}(S_t(C), F_t(C))$ and pick $C_t= \argmax_{C \in \mathcal{K}} \theta_C $
    \STATE For each $a \in C_t$ sample $\theta_a \sim \text{Beta}(S_t(a), F_t(a))$.
    \STATE Play arm $a_t = \argmax_{a \in C_t} \theta_a$ and collect  reward $r_t$. 
    \STATE Update $S_{t+1}(a_t) = S_{t}(a_t) + r_t$ ,  $F_{t+1}(a_t) = F_t(a_t) + (1-r_t)$.
    \STATE Update $S_{t+1}(C_t) = S_{t}(C_t) + r_t$ and $F_{t+1}(C_t) = F_t(C_t) + (1-r_t)$.
\ENDFOR
\end{algorithmic}
\end{algorithm} 

Given a clustering of the arms into a set of clusters $\mathcal{K}$, we introduce a natural two-level bandit policy based on TS,  Algorithm \ref{alg:TSC}.  In addition to the belief for each arm $a$,  $\text{Beta}(S_t(a), F_t(a))$,  the learner also keeps a belief over possible expected rewards $\text{Beta}(S_t(C), F_t(C))$ for each cluster $C \in \mathcal{K} $.  At each $t$,  the learner first use TS to pick a cluster - that is,  it samples $\theta_C \sim \text{Beta}(S_t(C), F_t(C))$ for each cluster $C \in \mathcal{K}$ and then considers the cluster $C_t = \argmax_{C \in \mathcal{K}} \theta_C$.   The learner then samples $\theta_a \sim \text{Beta}(S_t(a), F_t(a))$ for each $a \in C_t$ and plays the arm $a_t = \argmax_{a \in C_t} \theta_a$.  Given a reward $r_t$ the learner updates  the beliefs for $a_t$ and $C_t$ as follows $S_{t+1}(a_t) = S_{t}(a_t) + r_t$ ,  $F_{t+1}(a_t) = F_t(a_t) + (1-r_t)$,  $S_{t+1}(C_t) = S_{t}(C_t) + r_t$ and $F_{t+1}(C_t) = F_t(C_t) + (1-r_t)$.

We extended this two-level scheme to hierarchical clustering of depth L, by recursively applying TS at each level of the tree,  in Algorithm \ref{alg:HTS}.  The learner starts at the root of the hierarchical clustering, $\mathcal{T}$, and samples an expected reward for each of the sub-trees,  $\mathcal{T}_1^i$ spanned by its children, $i=1, ...$, from $\text{Beta}(S_t(\mathcal{T}_1^i), F_t(\mathcal{T}_1^i))$. The learner now traverses down to the root of the sub-tree satisfying $\mathcal{T}_{1, t}^i = \argmax_{\mathcal{T}_1^i} \theta_{\mathcal{T}_1^i}$. This scheme is recursively applied until the learner reaches a leaf,  i.e. an arm $a_t$, which is played.  Given a reward $r_t$, each belief along the path from the root to $a_t$ is updated using a standard TS update.

Algorithm \ref{alg:TSC} and \ref{alg:HTS} are not restricted to Bernoulli distributed rewards and can be used for any reward distribution with support $[0, 1]$ or for unbounded rewards by using Gaussian prior and likelihood in TS, as done for the standard MAB in \citet{Agrawal2017}.
\begin{algorithm}[t]
\caption{HTS}\label{alg:HTS}
\begin{algorithmic}
\REQUIRE $\mathcal{A}$, $\mathcal{T}$
\STATE Set $S_0(\mathcal{T}_l^i)=F_0(\mathcal{T}_l^i)=1$ for each sub-tree $\mathcal{T}_l^i$. 
\FOR{$t=1, ..., T$}
\STATE Set $\mathcal{T}_{t} = \mathcal{T}$.
\WHILE{$\mathcal{T}_{t}$ is not a leaf}
\STATE For each sub-tree $\mathcal{T}_l^i$ spanned by the children of $\mathcal{T}_t$ sample $\theta_{\mathcal{T}_l^i} \sim \text{Beta}(S_t(\mathcal{T}_l^i),  F_t(\mathcal{T}_l^i))$. 
\STATE Set $\mathcal{T}_{t} = \argmax \theta_{\mathcal{T}_l^i}$.
\ENDWHILE
\STATE Play the arm $a_t$ corresponding to the leaf $\mathcal{T}_{t}$ and collect the reward $r_t$.
\STATE Perform a TS update on each $S_t(\mathcal{T}_l^i),  F_t(\mathcal{T}_{l}^i)$ on the path to $a_t$.
\ENDFOR
\end{algorithmic}
\end{algorithm}

\subsection{Regret Analysis TSC}\label{sec:regret_TSC}
Assume that we have a clustering of $N$ Bernoulli arms, into a set of clusters $\mathcal{K}$. For each arm $a$, let $\mu_a$ denote the expected reward and let $a^*$ be the unique optimal arm with expected reward $\mu^*$. We denote the cluster containing $a^*$ as $C^*$. We denote the expected regret for each $a$ as $\Delta_a := \mu^* - \mu_a$ and for each cluster $C \in \mathcal{K}$, we define $\overline{\mu}_C = \max_{a \in C} \mu_a$, $\underline{\mu}_C = \min_{a \in C} \mu_a$ and $\Delta_C  =  \mu^* - \overline{\mu}_C$.

For each cluster $C \in \mathcal{K}$ we define distance $d_C$ to the optimal cluster $C^*$  as $d_C = \min_{a \in C^* ,\hat{a} \in C} \mu_{a} - \mu_{\hat{a}}$ and the width $w_C$ as $w_C = \overline{\mu}_C - \underline{\mu}_C$, let $w^*$ denote the width of the optimal cluster.

 \begin{assumption}[Strong Dominance]
  For $C \neq C^*, d_C>0$. 
 \end{assumption}
 This assumption is equivalent to what is referred to as \emph{tight clustering} in \citet{Bouneffouf} and \emph{strong dominance} in \citet{Matthieu2019}. In words, we assume that, in expectation, every arm in the optimal cluster is better than every arm in any suboptimal cluster. 
 
In order to bound the regret of TSC we will repeatedly use the following seminal result for the standard MAB case (without clustering) from \citet{Kaufmann12}. 
Here, we denote the Kullback-Leibler divergence between two Bernoulli distributions with means $\mu_1$ and $\mu_2$ as $\KL(\mu_1, \mu_2)$ and the natural logarithm of $T$ as $\log T$. 

\begin{theorem}[\citep{Kaufmann12}]
\label{th:kaufmann}
In the standard multi-arm bandit case with optimal arm reward $\mu^*$, the number of plays of a sub--optimal arm $a$ using TS is bounded from above, for any $\epsilon >0$, by
\[ (1+ \epsilon) \frac{1}{\KL(\mu_a, \mu^*)} (\log T + \log \log T)  + O(1). \]
\end{theorem}

Our plan is to apply Theorem \ref{th:kaufmann} in two different cases: to bound the number of times a  sub-optimal cluster is played and to bound the number of plays of a sub-optimal arm in the optimal cluster. However, the theorem not directly applicable to the number of plays of a sub-optimal cluster, $N_{C,T}$, since the reward distribution is drifting as the policy is learning about the arms within $C$. Nevertheless, we can use a comparison argument to bound the number of plays of a sub-optimal cluster by plays in an auxiliary problem with stationary reward distributions and get the following lemma.

\begin{lemma}\label{lm:subopt_clusters}
For any $\epsilon > 0$ and assuming strong dominance, the expected number of plays of a sub-optimal cluster $C$ at time $T$ using TSC is bounded from above by \begin{align*}
   E[N_{C, T}] \leq \frac{1 + \epsilon}{\KL(\overline{\mu}_C, \underline{\mu}_{C^*})} (\log T + \log \log T) + O(1).
\end{align*}
\end{lemma}

\iftoggle{proofs}{
    \begin{proof}
    Theorem \ref{th:kaufmann} is not directly applicable to the number of plays of sub-optimal clusters since the mean reward $\mu_C$ is drifting as the policy within cluster $C$ is learning. However, we argue that the number of sub-optimal plays is upper bounded by the number of plays in an auxiliary problem where the reward distribution is stationary.

W.l.o.g assume we have two clusters $c_1$ and $c_2$ where $c_2$ is the sub-optimal cluster. Let $\pi_1$ denote the TSC policy and $C_t$ the cluster played at time $t$. We have \begin{align*}
    P_{\pi_1}(C_t=c_2) \leq P_{\pi_2}(C_t=c_2)
\end{align*}
where $\pi_2$ uses Thompson sampling to pick cluster but will always play the best action in $c_2$ and the worst in $c_1$. To see this we make a simple induction argument. For $t=1$ we have  $P_{\pi}(C_1=c_2) = P_{\pi_2}(C_1=c_2)$, under the assumption that both policies starts with the same prior for the cluster part. For $t=2$ we have \begin{align*}
    & P_{\pi}(C_2 = c_2) = 
    \sum_{j, r}  P_{\pi}(C_2 = c_2| C_1 = c_j , r_t = r)  \\ & P_{\pi}(r_t = r |C_1 = c_j) P_{\pi}(C_1 = c_j)
\end{align*}
and by the definition of $\pi_2$ we know that it will always upper bound the probability of reward on $C_2$ for $\pi$ and lower bound it on $C_1$. Since both $\pi_1$ and $\pi_2$ uses Thompson sampling to pick cluster their posteriors on the cluster level will be the same conditioned on the same history. Thus, we have \begin{align*}
    P_{\pi_1}(C_2=c_2) \leq P_{\pi_2}(C_2=c_2).
\end{align*}

Now let $H_{t-1}$ denote the history up to time $t$ and assume $P_{\pi_1}(C_{t-1}=C_2) \leq P_{\pi_2}(C_{t-1}=C_2)$.

By the property of Thompson sampling we have for $i \in \{1, 2\}$ \begin{align*}
   & P_{\pi_i}(C_t = c_2 | r_{t-1}=1, C_{t-1}=c_2, H_{t-1}) \geq \\
   & P_{\pi_i}(C_t = c_2 | r_{t-1}=0, C_{t-1}=c_2, H_{t-1}) \\ 
\end{align*}
and 
\begin{align*}
   & P_{\pi_i}(C_t = c_2 | r_{t-1}=1, C_{t-1}=c_1, H_{t-1}) \leq \\
   & P_{\pi_i}(C_t = c_2 | r_{t-1}=0, C_{t-1}=c_1, H_{t-1}).
\end{align*}
Using this together with the induction assumption and the properties of $\pi_2$ we get
\begin{align*}
    & P_{\pi_1}(C_t = c_2) = \sum_{H_{t-1}}P_{\pi_1}(C_t = c_2 | H_{t-1}) p_{\pi_1}(H_{t-1}) = \\
    & \sum_{r, j, H_{t-2}}P_{\pi_1}(C_t = c_2 | r_{t-1}=r, C_{t-1}=c_j, H_{t-2}) p_{\pi_1}(H_{t-2}) \\
    & \cdot P_{\pi_1}(r_{t-1}=r | C_{t-1} = c_j, H_{t-2}) P_{\pi_1}(C_{t-1} = c_j|H_{t-2}) \\
   & \leq \sum_{H_{t-1}}P_{\pi}(C_t = c_2| H_{t-1})P_{\pi_2}(H_{t-1}) =  P_{\pi_2}(C_t = c_2).
\end{align*}

In the above inequality we have used the fact that conditioned on the same history, both $\pi_1$ and $\pi_2$ will have the same probability to pick a certain cluster, so we can change \begin{align*}
    P_{\pi_1}(C_t = c_2 | r_{t-1}=r, C_{t-1}=c_j, H_{t-2})
\end{align*}
to \begin{align*}
    P_{\pi_2}(C_t = c_2 | r_{t-1}=r, C_{t-1}=c_j, H_{t-2})
\end{align*}
and by construction of $\pi_2$ we know that \begin{align*}
    P_{\pi_2}(r_{t-1}=1 | C_{t-1}=c_2) \geq P_{\pi_1}(r_{t-1}=1 | C_{t-1}=c_2)
\end{align*}
and \begin{align*}
    P_{\pi_2}(r_{t-1}=0 | C_{t-1}=c_1) \geq P_{\pi_1}(r_{t-1}=0 | C_{t-1}=c_1)
\end{align*} since $\pi_2$ always plays the best arm in $c_2$ and worst arm in $c_1$. By the induction assumption we have \begin{align*}
   & P_{\pi_1}(C_{t-1}=c_2) = \sum_{H_{t-2}} P_{\pi_1}(C_{t-1}= c_2 | H_{t-2})P_{\pi_1}(H_{t-2}) \\ 
   & \leq \sum_{H_{t-2}} P_{\pi_2}(C_{t-1}= c_2 | H_{t-2})P_{\pi_2}(H_{t-2}) = P_{\pi_2}(C_{t-1}=c_2)
\end{align*} 
Combining these facts yields the last step in the inequality.
Since $P_{\pi_1}(C_t=c_2) \leq P_{\pi_2}(C_t=c_2)$ we know that \begin{align*}
    \E_{\pi_1}[N_{C_2,T}] \leq \E_{\pi_2}[N_{C_2, T}]
\end{align*}
and we can apply Theorem \ref{th:kaufmann} to upper bound the expectation to the right.
    \end{proof}
}

We can use Lemma \ref{lm:subopt_clusters} to derive the following instance-dependent regret bound for TSC.

\begin{theorem}\label{th:instance_dependent} For any $\epsilon > 0$, the expected regret of TSC under the assumption of strong dominance is bounded from above by\begin{align*}
    & (1+\epsilon)  \left(\sum_{C\neq C^*} \frac{\Delta_C}{\KL(\overline{\mu}_C, \underline{\mu}_{C^*})} +  \sum_{a \in C^*} \frac{\Delta_a}{\KL(\mu_a, \mu^*)}\right) \log T  +  \\
    & + o(\log T).
\end{align*}
\end{theorem}

\iftoggle{proofs}{
\begin{proof}
We can decompose the regret into \begin{align*}
    E[R_T] = \sum_{C\neq C^*}\sum_{a \in C} \Delta_a \E[N_{a, T}] + \sum_{a \in C^*} \Delta_a \E[N_{a, T}]
\end{align*}
where the first term consider the regret suffered from playing sub-optimal clusters and the second term regret suffered from playing sub-optimal arms within the optimal cluster. The second term can be bounded by just applying Theorem \ref{th:kaufmann} for $\epsilon > 0$ \begin{align*}
    \sum_{a \in C^*} \Delta_a \E[N_{a, T}] \leq (1 + \epsilon) \sum_{a \in C^*} \frac{1}{\Delta_a} \log T + o(\log T).
\end{align*}

To bound the first term, consider sub-optimal cluster $C$ and let $N_{C, T}$ denote the number of times we play $C$. Let $a_C^*$ be the action with highest expected reward in $C$. Then for any other $a \in C, \, a\neq a_C^*$ we can bound the number of plays, $N_{a, T_{C, T}}$, by Theorem \ref{th:kaufmann} \begin{align*}
     &\E[N_{a_C, N_{C, T}}] \leq (1 + \epsilon) \frac{1}{\KL(\mu_a, \mu_{a^*})}(\log N_{C, T} + \log \log N_{C, T}) \\ 
     & + O(1)
\end{align*}
and for $a_C^*$ we have 
\begin{align*}
    \E[N_{a_C^*, N_{C, T}}] \leq \E[N_{C, T}].
\end{align*}
From Lemma \ref{lm:subopt_clusters} we know that for $\epsilon > 0$ \begin{align*}
    \E[N_{C, T}] \leq (1 + \epsilon)\frac{1}{\KL(\overline{\mu}_C, \underline{\mu}_{C^*})}(\log T + \log \log T) + O(1)
\end{align*}

and we thus get a $\log \log T$ dependence on all arms in $C$ except the one with highest expected reward \begin{align*}
    & \E[N_{a, N_{C, T}}] \leq (1 + \epsilon)\frac{1}{\KL(\mu_a, \mu_{a^*})} \log \log T + o(\log \log T) \\
    &  \E[N_{a_C^*, N_{C, T}}] \leq (1 + \epsilon)\frac{1}{\KL(\overline{\mu}_C, \underline{\mu}_{C^*})}\log T + o(\log T).
\end{align*}

Therefore we can bound the regret suffered from sub-optimal clusters for any $\epsilon > 0 $ as \begin{align*}
    &\sum_{C\neq C^*}\sum_{a \in C} \Delta_a \E[N_{a, T}]\\
    &\leq (1 + \epsilon)(\sum_{C\neq C^*} \frac{\Delta_C }{\KL(\overline{\mu}_C, \underline{\mu}_{C^*})}\log T  + \\
    & + \sum_{a \in C, a\neq a^*} \frac{\Delta_a}{\KL(\mu_a, \mu_{a^*})}\log \log T)  + o(\log T) \\
    & \leq (1 + \epsilon)\sum_{C\neq C^*} \frac{\Delta_C }{\KL(\overline{\mu}_C, \underline{\mu}_{C^*})}\log T + o(\log T). 
\end{align*}

Combining with the bound on regret within the optimal cluster $C^*$ yields the instance-dependent regret bound \begin{align*}
    & \E[R_T] \leq \\
    & \leq (1+\epsilon)\left(\sum_{C\neq C^*} \frac{\Delta_C}{\KL(\overline{\mu}_C, \underline{\mu}_{C^*})}  +  \sum_{a \in C^*} \frac{\Delta_a}{\KL(\mu_a, \mu^*)}\right) \log T \\
    &+ o(\log T).
\end{align*}
\end{proof}
}

We can derive an instance-independent upper bound from Theorem~\ref{th:instance_dependent} which only depends on number of clusters, number of arms in the optimal cluster and the quality of the clustering . Now, define $\gamma_C$ as the ratio between width of the optimal cluster and the distance of $C$ to the optimal cluster:
\[
\gamma_C :=  
\begin{cases}
    w^*/d_C, & C \neq C^* \\
    0, & \text{otherwise}
\end{cases}
\]
and let $\gamma := \sum_{C} \gamma_C/K$. We arrive at the following result.

\begin{theorem}\label{th:instance_independent}
Assume strong dominance and let $A^*$ be the number of arms in the optimal cluster and $K$ the number of sub-optimal clusters. The expected regret of TSC is  bounded from above by $\E[R_T] \leq O\left(\sqrt{(A^* + K(1 + \gamma))T \log T}\right)$.
\end{theorem}

 \iftoggle{proofs}{
\begin{proof}
We rewrite $\Delta_C = d_C + w^*$  where $w^*$ is the width of the optimal cluster and hence by the definition of $\gamma_C$ we have \begin{align*}
    \Delta_C = (1 + \gamma_C) d_C.
\end{align*}
By Pinsker's inequality we have \begin{align*}
    \KL(\overline{\mu}_C, \underline{\mu}_{C^*}) \geq 2 d_C^2
\end{align*}
and for arms in the optimal cluster we have \begin{align*}
    \KL(\mu_a, \mu^*) \geq 2 \Delta_a^2
\end{align*}

Thus, the instance-dependent regret bound can be upper-bounded by \begin{align*}
 \frac{1+\epsilon}{2}\left(\sum_{C\neq C^*} \frac{1 + \gamma_C}{d_C}  +  \sum_{a \in C^*} \frac{1}{ \Delta_a}\right) \log T + o(\log T). 
\end{align*}

Let $\Delta > 0$. 
\begin{itemize}
    \item For all clusters $C$ and arms $a \in C^*$ such that $d_C, \Delta_a < \Delta$, the cumulative regret from these are upper-bounded by $\Delta T$. 
    \item For each cluster $C$ such that $d_C \geq \Delta$ the amount of regret suffered from playing $C$ is $O(\frac{1 + \gamma_C}{\Delta}\log T)$ and for each $a \in C^*$ the regret suffered is $O(\frac{1}{\Delta} \log T)$. In total this is $O(\frac{A^* + K(1 + \gamma)}{\Delta} \log T)$. 
\end{itemize}

Combining this yields \begin{align*}
    \E[R_T] \leq O(\Delta T + \frac{A^* + K(1 + \gamma) }{\Delta} \log T).
\end{align*}
Since this holds $\forall \Delta > 0$ we pick $\Delta = \sqrt{\frac{(A + K(1 + \gamma))\log T}{T}}$ and hence, \begin{align*}
    \E[R_T] \leq O\left(\sqrt{(A^* + K(1 + \gamma))T \log T}\right).
\end{align*}
\end{proof}
}

\paragraph{Clustering Quality and Regret.} As a sanity check, we note that if the expected rewards of all arms in the optimal cluster are equal we have $\gamma=0$ and the bound in Theorem \ref{th:instance_independent} reduces to the bound for the standard MAB in \cite{Agrawal2017} with $K+1$ arms. On the other hand, if the optimal cluster has a large width along with many sub-optimal clusters with a small distance to the optimal cluster $\gamma$ becomes large and little is gained from the clustering. Two standard measures of cluster quality are the (a) the maximum diameter/width of a cluster and (b) inter-cluster separation. We see that for our upper bound, \emph{only the width of the optimal cluster and the separation of other clusters from the optimal cluster} are important. These dependencies are consistent with the observations in \citet[Section 5.3]{Pandey2007}, which suggest that high cohesiveness within the optimal cluster and large separation are crucial for achieving low regret. However our analysis is more precise than their observations and we also provide rigorous regret bounds.

\subsection{Lower Bounds for Disjoint Clustering}\label{sec:lower_bounds}
 In the case of Bernoulli distributed rewards we can derive the following lower bound for the instance dependent case using the pioneering works of \citet{Lai1985}.
 
 \begin{theorem}\label{th:lower_bound}
 The expected regret for any policy, on the class of bandit problems with Bernoulli distributed arms clustered such that strong dominance holds, is bounded from below by \begin{align*}
     \lim_{T \xrightarrow{} \infty} \inf \frac{\E[R_T]}{\log T} \geq \sum_{a \in C^*} \frac{\Delta_a}{\KL(\mu_a, \mu^*)} + \sum_{C \neq C^*} \frac{\Delta_C }{\KL(\underline{\mu}_C, \mu^*)}
 \end{align*}
 \end{theorem}
 
 \iftoggle{proofs}{
 \begin{proof}
 We make use of the pioneering work of \cite{Lai1985} which gives that \begin{align}\label{eq:Lai}
     \lim_{T \xrightarrow{} \infty} \inf \frac{\E[R_T]}{\log T} \geq \sum_{a} \frac{\Delta_a}{\KL(\mu_a, \mu^*)}
 \end{align}
 for a standard multi-armed bandit with Bernoulli rewards. We can decompose the regret over sub-optimal clusters and sub-optimal arms in the optimal cluster \begin{align*}
     \E[R_T] = \sum_{C\neq C^*} \sum_{a \in C} \Delta_a \E[N_{a,T}] + \sum_{a \in C^*}\Delta_a \E[N_{a,T}], 
 \end{align*}
 and using the fact that the regret suffered within a sub-optimal cluster is bounded from below by the smallest regret in the cluster \begin{align*}
    \sum_{a \in C} \Delta_a \E[N_{a,T}]  \geq \Delta_C \sum_{a \in C} \E[N_{a, T}].
 \end{align*}
 Now we get the proposed bound by independently bounding each term from below by Equation:\ref{eq:Lai} and using the fact that for any cluster $C$ and any arm $a \in C$ we have \begin{align*}
     \KL(\mu_a, \mu^*) \geq \KL(\underline{\mu}_C, \mu^*).
 \end{align*}
 \end{proof}
 }
 
We compare the lower bound in Theorem \ref{th:lower_bound} to our instance-dependent upper bound in Theorem \ref{th:instance_dependent} and we see that the regret suffered in TSC from playing sub-optimal clusters asymptotically differs from the corresponding term in the lower bound  by a factor depending on the width of the clusters since \begin{align*}
     \KL(\overline{\mu}_C, \underline{\mu}_{C^*}) = \KL(\underline{\mu}_C + w_C, \mu^* - w^*) \leq \KL(\underline{\mu}_C, \mu^*).
 \end{align*}
Thus, as the width of the clusters goes to zero, the regret of TSC approaches the lower bound. However, as also discussed in \citet{Matthieu2019} it is unclear whether the lower bound derived in Theorem \ref{th:lower_bound} can be matched by any algorithm since it doesn't depend at all on the quality on the given clustering and assumes the optimal policy to always play the worst action in sub-optimal clusters.

The following minimax lower bound follows trivially from the $\Omega(\sqrt{NT})$ minimax lower bound for standard MAB \citep{Auer1998} by considering the two cases: where all clusters are singletons and all arms are in one cluster.  
 
 \begin{theorem}\label{th:independent_lower}
 Let $K$ be the number of sub-optimal clusters and let $A^*$ be the number of arms in the optimal cluster. The expected regret for any  policy, on the class of bandit problems with Bernoulli distributed arms clustered such that strong dominance holds,  satisfies $\E[R_T] \geq \Omega(\sqrt{(A^* + K)T})$.
 \end{theorem}
 
 \iftoggle{proofs}{
 \begin{proof}
 First consider the case where all arms are assigned to the same cluster. Any algorithm needs to at least have a $\sqrt{A^*T}$ dependence in the regret otherwise the lower bound $\Omega(\sqrt{NT})$ would be violated. 
 
 Secondly, consider the case where all clusters only contain one arm each. We have that any algorithm needs at least a $\sqrt{KT}$ dependence otherwise  $\Omega(\sqrt{NT})$ would be violated. 
 
 Since $\sqrt{K + A^*} \leq \sqrt{K} + \sqrt{A^*}$ it follows that for any algorithm we have \begin{align*}
     \E[R_T] \geq \Omega(\sqrt{(A^* + K)T}).
 \end{align*}
 \end{proof}
 }

Let $d>0$ be the smallest distance between any sub-optimal and the optimal cluster. We compare Theorem \ref{th:independent_lower} to the upper bound in Theorem \ref{th:instance_independent} and observe that $\sqrt{(A^* + K)T} \leq \sqrt{(A^* + (1+\gamma)K)T}  \leq \sqrt{\left(1 + \frac{1}{d}\right)} \sqrt{(A^* + K)T}$.
Hence, our upper bound in Theorem \ref{th:instance_independent} matches the lower bound up to logarithmic factors and a constant depending on the separation of the clusters. 

\subsection{Regret Analysis HTS}\label{sec:regret_HTS}
Assume we have $N$ Bernoulli arms clustered into a tree $\mathcal{T}$ and for simplicity we assume it to be perfectly height-balanced. We denote the sub-tree corresponding to node $j$ on level $i$ as $\mathcal{T}_{i}^{j}$ and on each level $i$ we denote the sub-tree containing the optimal arm as $\mathcal{T}_{i}^*$. Let $\mathcal{T}_{i+1}^{j}, \, j \in [1, K_{i}^*]$, denote sub-trees spanned by the child nodes of the root in $\mathcal{T}_{i}^*$, where $K_i^*$ is the number of children of the root in $\mathcal{T}_{i}^*$.  W.l.o.g let $j=1$ be the sub-tree, $\mathcal{T}_{i+1}^{1}$, that contains the optimal action. For each sub-tree $\mathcal{T}_{i}^{j}$ we define $
    \Delta_{i}^{j} := \mu^* - \max_{a \in \mathcal{T}_{i}^{j}} \mu_a$ and $
    d_{j}^{i } := \min_{a \in \mathcal{T}_{i}^{*}} \mu_a - \max_{a \in \mathcal{T}_{i}^{j}} \mu_a$.

 \begin{assumption}[Hierarchical Strong Dominance]
  We  assume $d_{i}^{j}>0, \, \forall i,j$  except for $\mathcal{T}_{i}^{*}$. 
 \end{assumption}
Under this assumption the results in  Theorem \ref{th:instance_dependent} can be naturally extended to HTS by recursively applying Theorem \ref{th:instance_dependent}.
 
 \begin{theorem}\label{thm:HTS}
 Assuming hierarchical strong dominance. For any $\epsilon > 0$, the expected regret of HTS is upper bounded by \begin{align*}
     (1 + \epsilon) \left ( \sum_{i=0}^{L-1} \sum_{j=2}^{K_{i}^*} \frac{\Delta_{i}^{j}}{(d_{j}^{i})^2} + \sum_{a \in \mathcal{T}_{L}^*} \frac{1}{\Delta_a} \right) \log T + o(\log T).
 \end{align*}
 \end{theorem}

 \iftoggle{proofs}{
 \begin{proof}
  We decompose the cumulative regret into \begin{align*}
     R_T := \sum_{\mathcal{T}_{1}^j \neq \mathcal{T}_{1}^*}\sum_{a \in \mathcal{T}_{1}^j} \Delta_a \E[N_{a,T}] + \sum_{a \in \mathcal{T}_{1}^*} \Delta_a \E[N_{a,T}].
 \end{align*}
 Since strong dominance holds on each level we bound the first sum by $\sum_{j=2}^{K_{0}^{*}}\frac{\Delta_{1}^j}{(2d_{1}^j)^2}\log T + o(\log T)$ using Theorem \ref{th:instance_dependent}, where $(2d_{1}^j)^2$ follows from Pinsker's inequality for Bernoulli distributions. We are left with bounding the regret from \begin{align*}
     \sum_{a \in \mathcal{T}_{1}^*} \Delta_a \E[N_{a,T}] = \sum_{j=2}^{K_1^*}\sum_{a \in \mathcal{T}_{2}^j} \Delta_a \E[N_{a,T}] + \sum_{a \in \mathcal{T}_{2}^*} \Delta_a \E[N_{a,T}].
 \end{align*}
 And we recursively apply Theorem \ref{th:instance_dependent} to bound the first time like above, until we reach level $L$ for which we use Theorem \ref{th:kaufmann} along with Pinsker's inequality to get \begin{align*}
    \sum_{a \in \mathcal{T}_{L}^*} \Delta_a \E[N_{a,T}] \leq (1 + \epsilon) \sum \frac{1}{\Delta_a} \log T + o(\log T)
 \end{align*}
 \end{proof}
 }

For $L=0$ Theorem \ref{sec:regret_HTS} reduces to the instance-dependent bound for standard TS and for $L=1$ it reduces to the bound for TSC presented in Theorem \ref{th:instance_dependent}. Hierarchical structures and bandits have previously been studied in the prominent works \citet{Coquelin07} and \citet{Bubeck11} which assumes there is a known smoothness. Here we do not make such assumptions and Theorem \ref{thm:HTS} instead relies on an assumption regarding the ordering of the tree.

\paragraph{Plausibility of Hierarchical Strong Dominance.} The hierarchical strong dominance assumption is perhaps too strong for a general hierarchical clustering but it might be reasonable for shallow trees. One example is in e-commerce where products can be organized into sub-categories and later categories. A user might have a strong preference for the sub-category ``Football'' in the category ``Sports''.  

\section{Contextual Bandit with Linear Rewards and Clustered Arms}\label{sec:CBC}
In this section, we consider the MABC problem in its contextual variant with linear expected rewards (CBC). As in the classic CB, there is for each arm $a \in \mathcal{A}$ an, a priori, unknown vector $\theta_a \in \mathbf{R}^d$. At each time $t$, the learner observes a context vector $x_t \in \mathbf{R}^d$ and the expected reward for each arm $a$ at time $t$, given that the learner has observed $x_t$, is 
$
    \E[r_t(a)|x_t] = x_{t}^\top \theta_a. 
$
Similar to MABC, the learner has, in addition, access to a clustering of the $N$ arms and for CBC we assume the arms to be clustered into a set of $\mathcal{K}$ disjoint clusters. 

For the CBC we extend TSC, Algorithm \ref{alg:TSC}, to LinTSC, as defined in Algorithm \ref{alg:LinTSC}. At each level of LinTSC, we use the Thompson sampling scheme developed for standard CB in \citet{Agrawal12b}. 
\begin{algorithm}
\caption{LinTSC}\label{alg:LinTSC}
\begin{algorithmic}
\REQUIRE $v>0$
\STATE Set $B_c=\mathbf{1}_d$, $f_c=\mathbf{0}_d$, $\mu_c=\mathbf{0}_d$, $B_{c,i}=\mathbf{1}_d$, $f_{c,i}=\mathbf{0}_d$, $\mu_{c,i} = \mathbf{0}_d$
\FOR{$t=1, ..., T$}
\STATE Observe context $x_t$
\STATE Sample $\theta_c \sim \mathcal{N}(\mu_c^\top x_t, v x_t^\top B_{c}^{-1}x_t)$
\STATE Consider cluster $k = \argmax_c \theta_c$ \
\STATE Sample $\theta_{k,i} \sim \mathcal{N}(\mu_{k,i}^\top x_t, v x_t^\top B_{k,i}^{-1}x_t)$
\STATE Play arm $a = \argmax_i \theta_{k, i}$ 
\STATE Observe reward $r_t$ and update $B_k = B_k + x_{t}x_{t}^\top$, $B_{k,a} = B_{k,a} + x_{t}x_{t}^\top$, $f_k = f_{k} + r x_t$, $f_{k,i} = f_{k,i} + r x_t$, $\mu_k = B_{k}^{-1}f_{k}$ and $\mu_{k,i} = B_{k,i}^{-1}f_{k,i}$ .
\ENDFOR
\end{algorithmic}
\end{algorithm}

\begin{figure*}%
\centering
\begin{tabular}{ccc}
\subfloat[$w^*=0.1$, $N=100$, $A^*=10$, $K=10$.]{\includegraphics[width=0.3\textwidth]{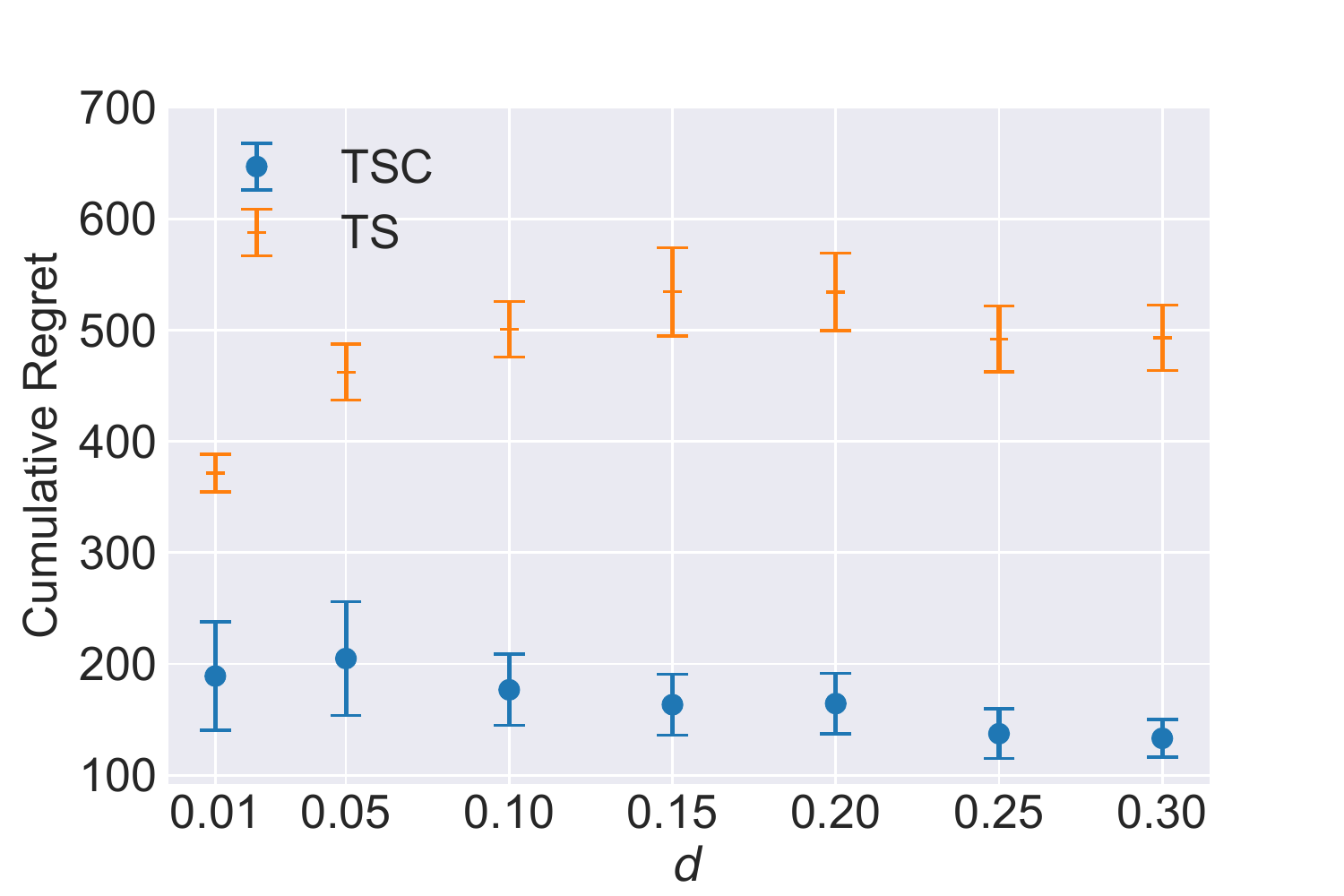}\label{fig:dependence_d}} &
\subfloat[$d=0.1$, $N=100$, $A^*=10$, $K=10$.]{\includegraphics[width=0.3\textwidth]{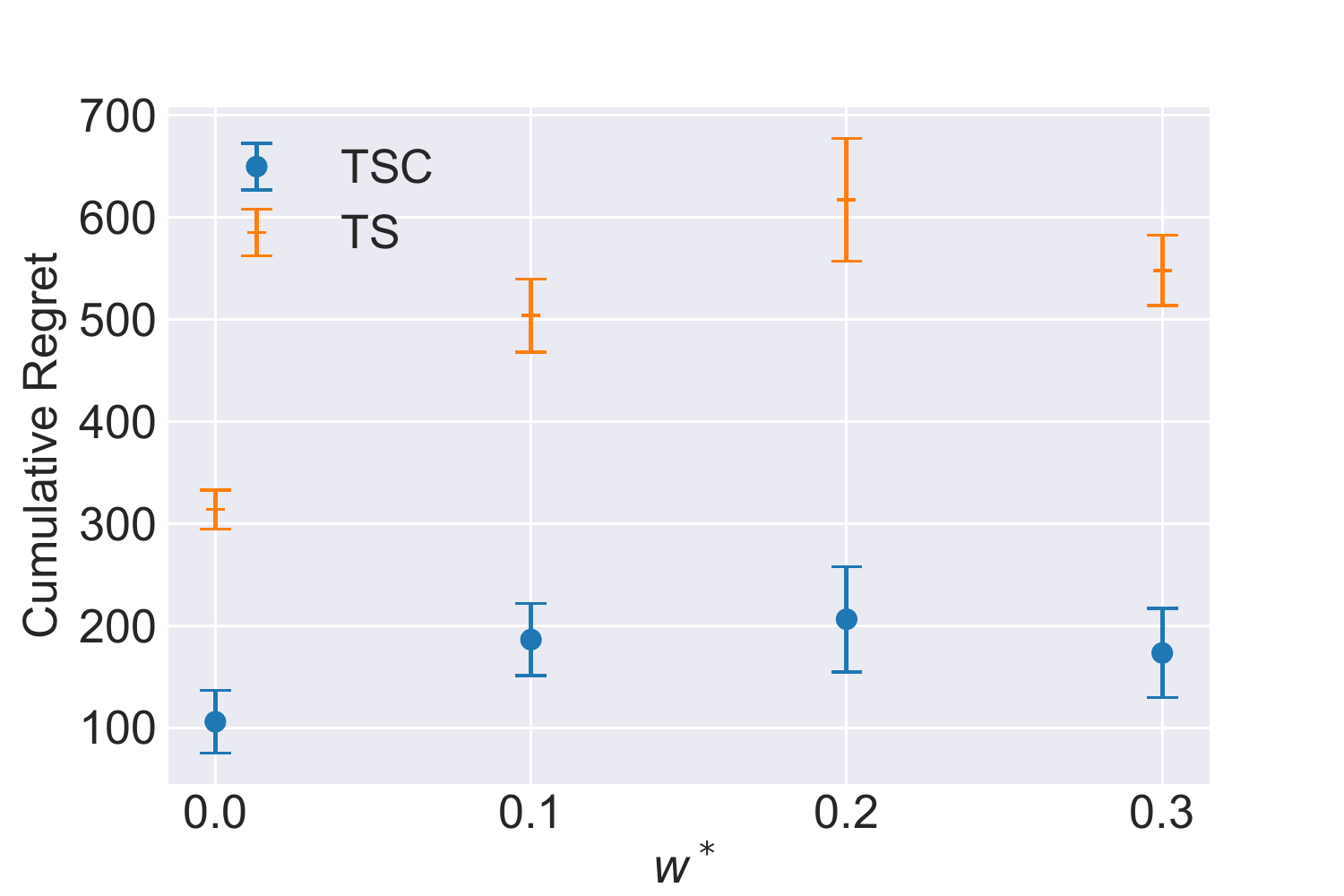}\label{fig:dependence_w}} & 
\subfloat[$w^*=0.1$, $d=0.1$, $A^*=K=\lfloor\sqrt{N}\rfloor$.]{\includegraphics[width=0.3\textwidth]{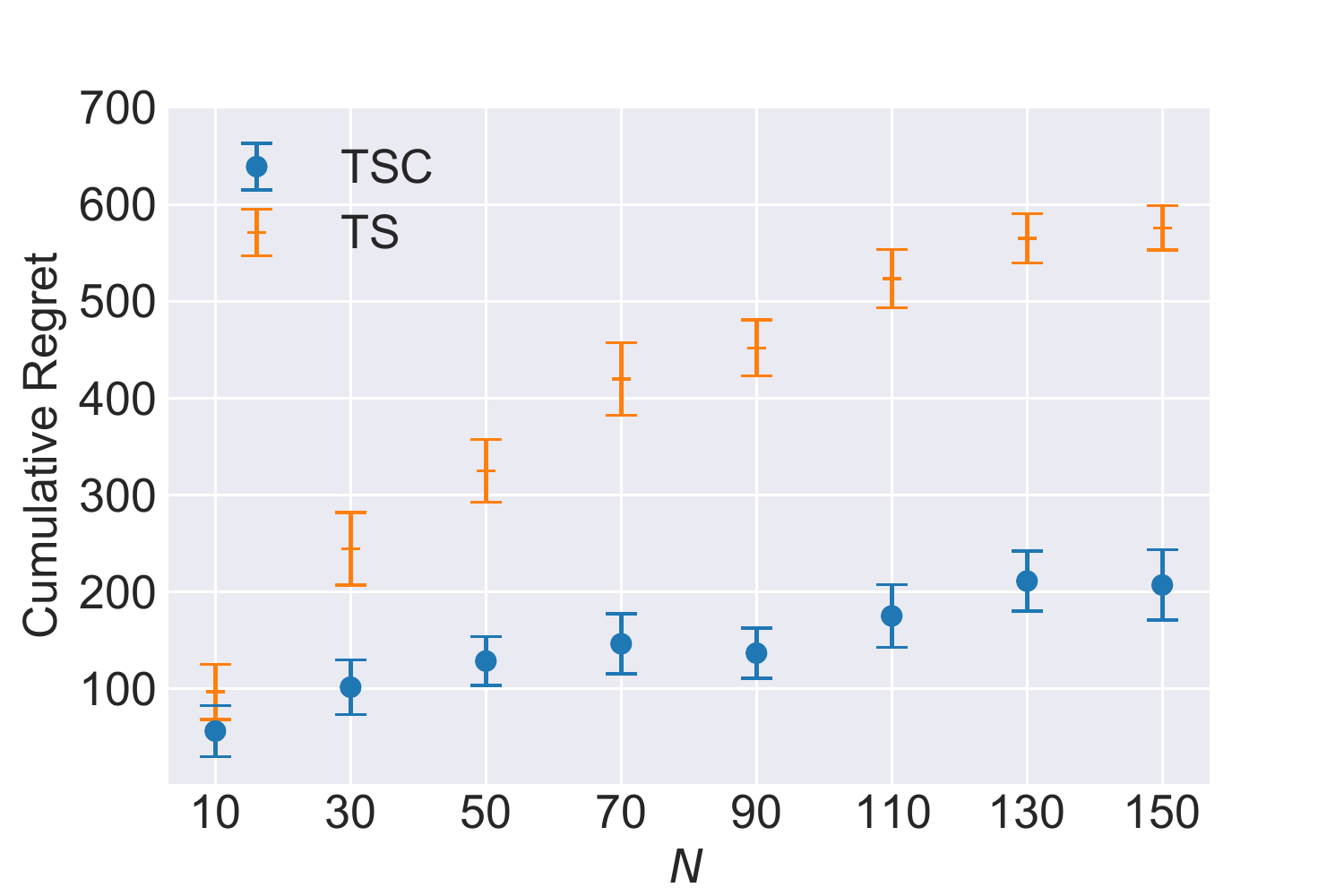}\label{fig:dependence_N}} \\
\subfloat[$w^*=d=0.1$, $N=100$, $A^*=10$.]{\includegraphics[width=0.3\textwidth]{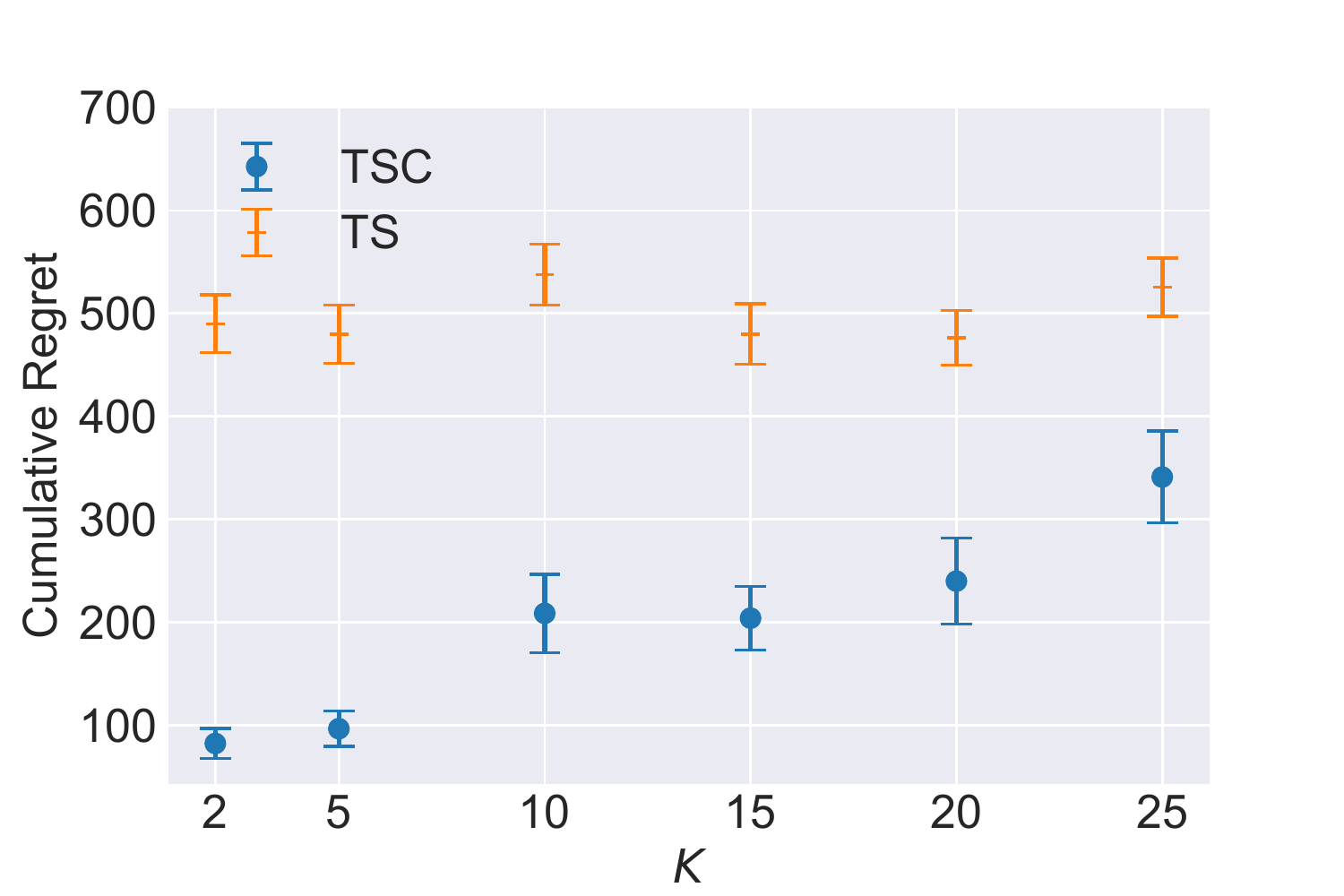}\label{fig:dependence_K}} &
\subfloat[$w^*=d=0.1$, $N=100$, $K=10$.]{\includegraphics[width=0.3\textwidth]{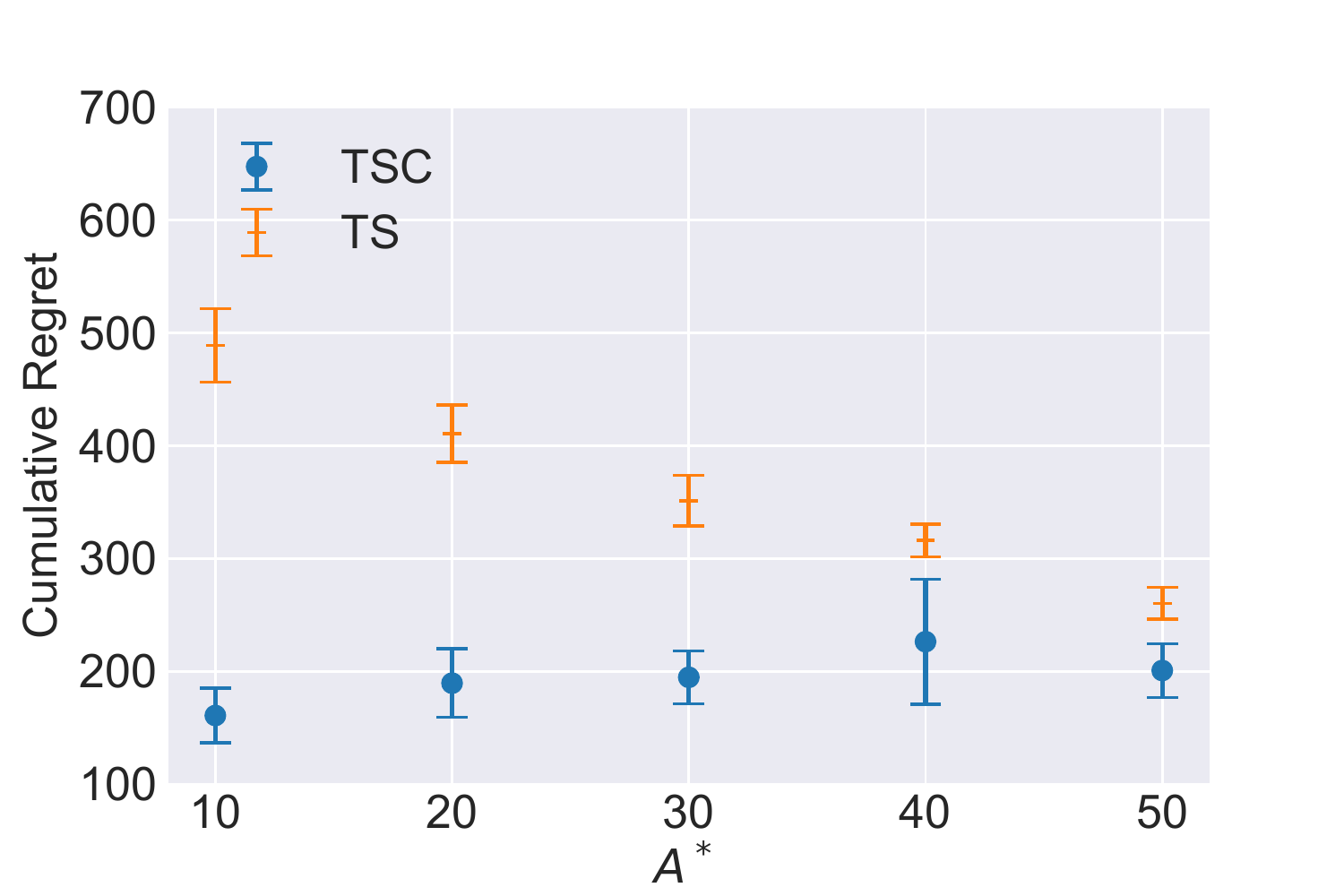}\label{fig:dependence_A}} &
\subfloat[Dependence on the number of levels $L$.]{\includegraphics[width=0.3\textwidth]{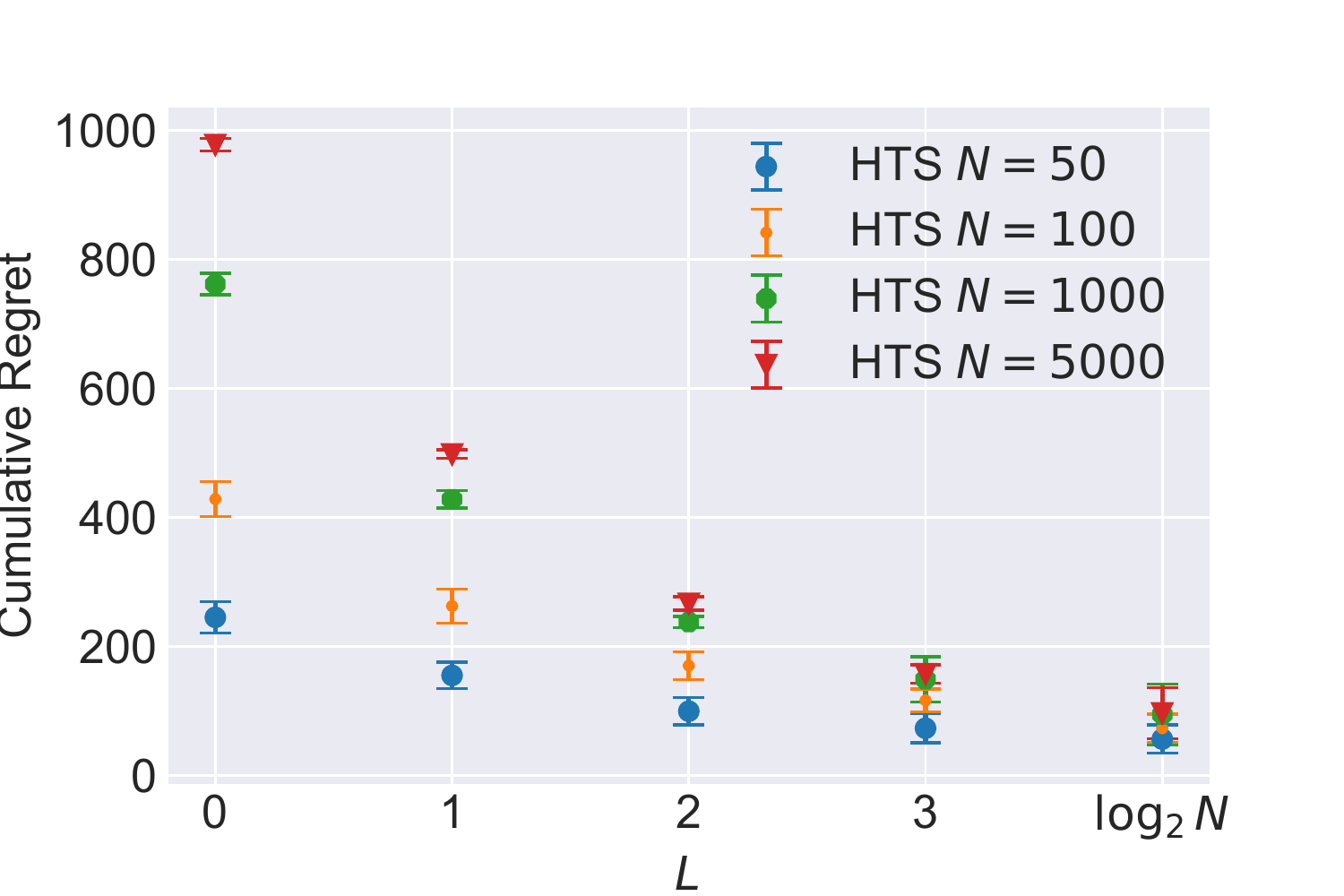}\label{fig:dependence_L}} 
\end{tabular}
\caption{Strong and Hierarchical Strong Dominance.}
\end{figure*}

\section{Experimental Results}\label{sec:results}
\subsection{Stochastic Multi-armed Bandit}
\paragraph{Strong Dominance.}
 We generate synthetic data, for which strong dominance holds, in the following way: We have $N$ arms and each arm $i$ is Bernoulli distributed with reward probability $p_i$. The arms are clustered into $K$ clusters and we have $A^*$ arms in the optimal cluster. For the remaining $N-A^*$ arms we assign each arm to one of the sub-optimal clusters with uniform probability.  We set the reward probability of the best arm in the optimal cluster to be $0.6$ and for the worst arm in the optimal cluster we set it to be $0.6-w^*$. For the remaining $A^*-2$ arms in the optimal cluster we draw the reward probability from $\mathcal{U}(0.6-w^*, 0.6)$ for each arm. In each sub-optimal cluster we set the probability of the best arm to be $0.6-w^*-d$ and for the worst arm to be $0.5-w^*-d$, the probability for the remaining arms are drawn from $\mathcal{U}(0.5-w^*-d, 0.6-w^*-d)$. The optimal cluster will then have a width of $w^*$ and the distance from each sub-optimal cluster to the optimal cluster will be $d$. In Figures~\ref{fig:dependence_d}--\ref{fig:dependence_A}, we run TS and TSC on the same instances for $T=3000$ time steps, varying the different instance parameters and plotting the cumulative regret of each algorithm at the final time step $T$. For each set of parameters we evaluate the algorithms using $50$ different random seeds and the error bars corresponds to $\pm 1$ standard deviation.  In Figures \ref{fig:dependence_d} and \ref{fig:dependence_w}, we observe that the cumulative regret scales depending on the clustering quality parameters $d$ and $w^*$ as suggested by our bounds in Section \ref{sec:regret_TSC}---that is, the cumulative regret of TSC decreases as $d$ increases and increases as $w^*$ increases. In Figure \ref{fig:dependence_N}, we observe that the linear dependence in $N$ for TS is changed to a linear dependence in $K$ and $A^*$, Figures \ref{fig:dependence_K} and \ref{fig:dependence_A}, which greatly reduces the regret of TSC compared to TS as the size of the problem instance increases. In Figure \ref{fig:dependence_A} we also see that as the number of arms in the optimal cluster, $A^*$, increases to be a substantial amount of the total number of arms, the gain from using TSC compared to TS vanishes. 

\paragraph{Hierarchical Strong Dominance.}
We generate a bandit problem by first uniformly sample $N$ Bernoulli arms from $\mathcal{U}(0.1, 0.8)$ followed by recursively sorting and merging the arms into a balanced binary tree, which has the hierarchical strong dominance property. In Figure \ref{fig:dependence_L}, we ran the algorithms for $T=3000$ over $50$ random seeds and illustrated how the cumulative regret at time $T$ of HTS changes as we alter the depth $L$ of the given tree and the total number of arms $N$. Note that $L=0$ corresponds to TS and $L=1$ corresponds to TSC. We observe that as the size of the problem instance grows, i.e increasing $N$, using more levels in the tree becomes more beneficial due to aggressive exploration scheme of HTS. Hence, once we realize that one sub-tree is better than the other we discard all arms in the corresponding sub-optimal sub-tree. Connecting back to Theorem \ref{thm:HTS} we see that HTS gets only a dependence $O(\log_2 N)$ in the number of arms when using the full hierarchical tree in Figure \ref{fig:dependence_L}.

\begin{figure*}%
\centering
\begin{tabular}{ccc}
\subfloat[K-means instance  with $N=100$, $K=10$.]{\includegraphics[width=0.3\textwidth]{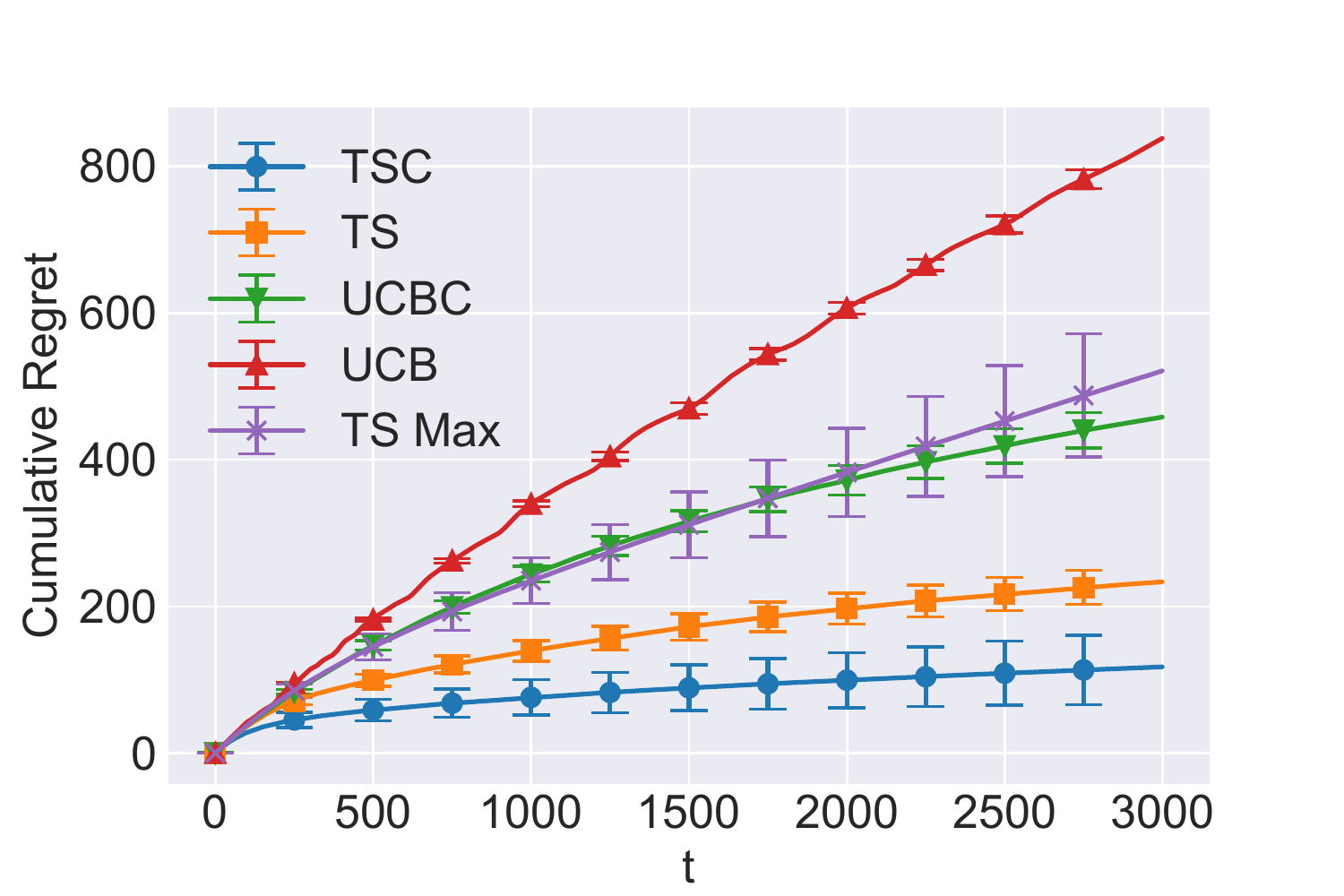}\label{fig:kmeans_small}}&
\subfloat[K-means instance  with $N=1000$, $K=32$.]{\includegraphics[width=0.3\textwidth]{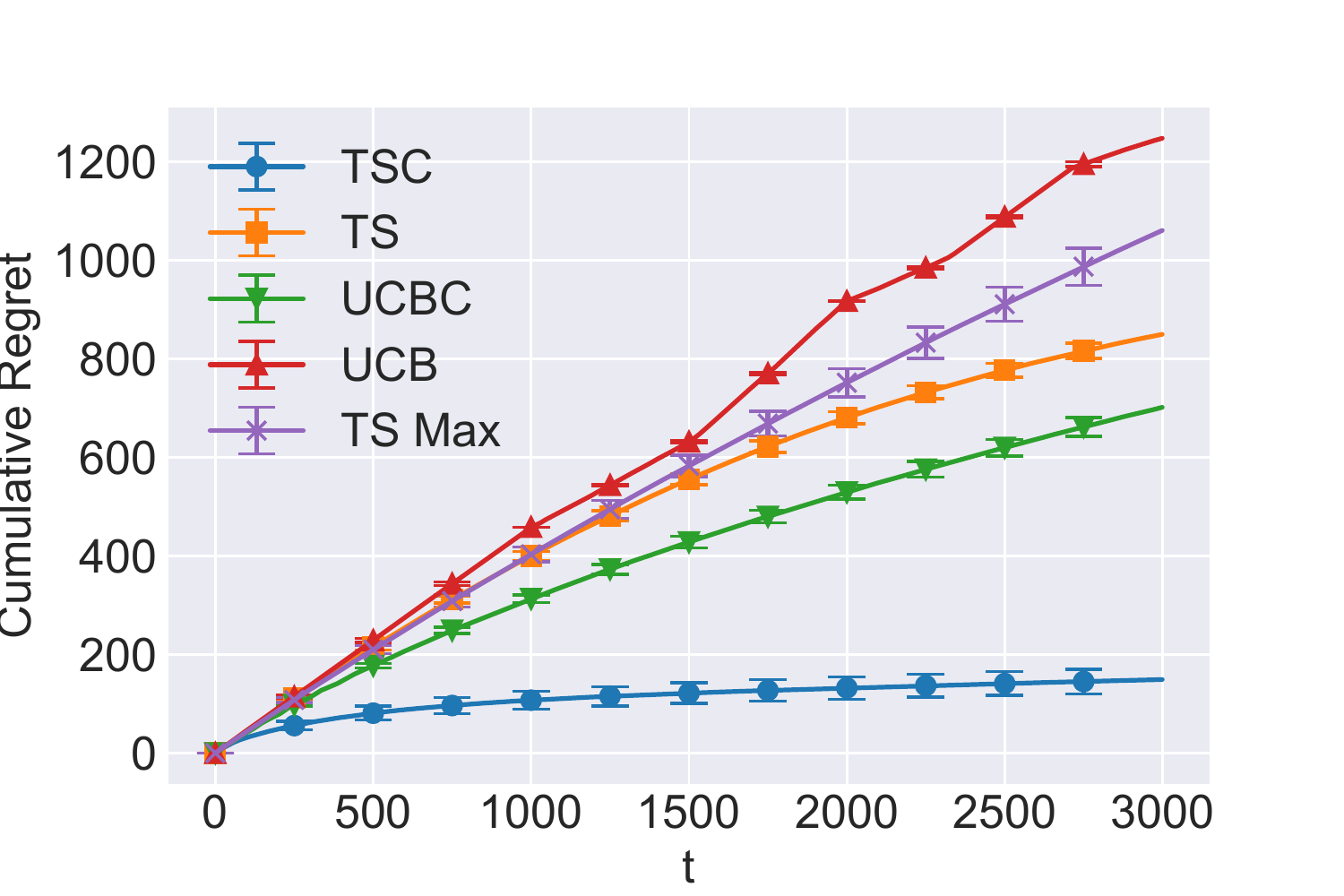}\label{fig:kmeans_large}} & 
\subfloat[Hierarchical clustering with k-means.]{\includegraphics[width=0.3\textwidth]{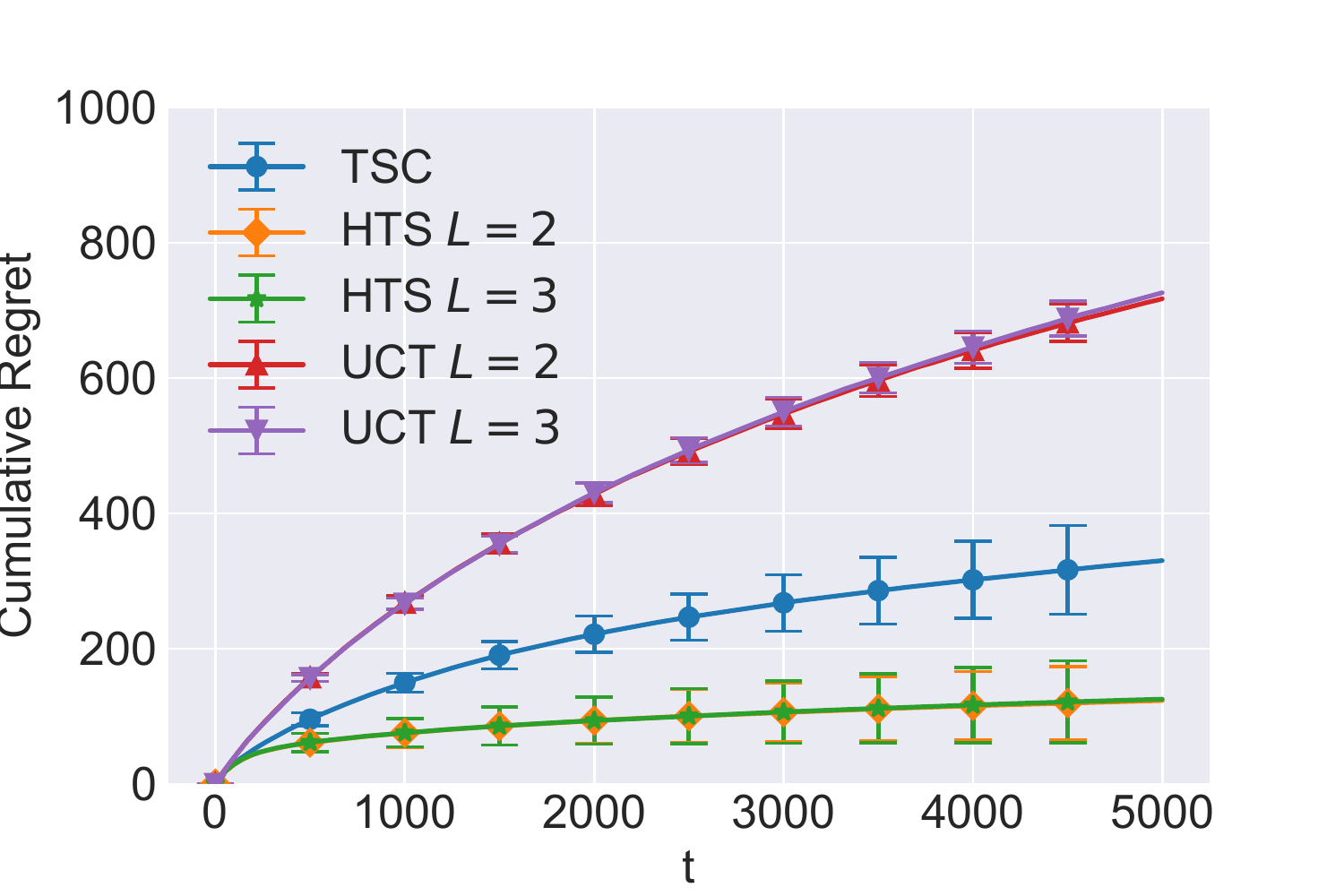}\label{fig:HTS_UCT}}
\\
\subfloat[CBC with $k=20$, $n=400$, $\epsilon=0.5$]{\includegraphics[width=0.3\textwidth]{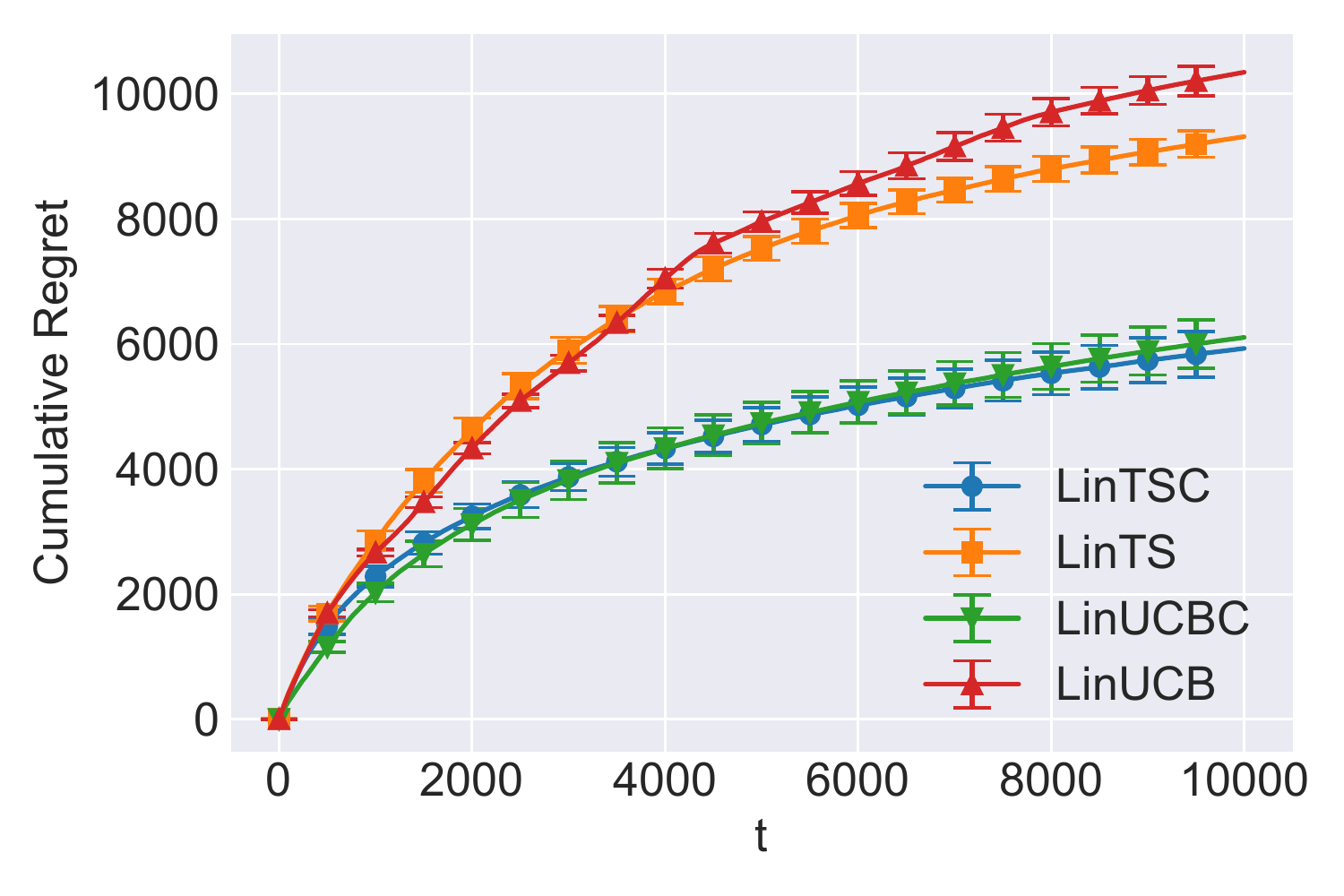}\label{fig:context_small}} &
\subfloat[CBC with $k=30$, $n=900$, $\epsilon=0.5$]{\includegraphics[width=0.3\textwidth]{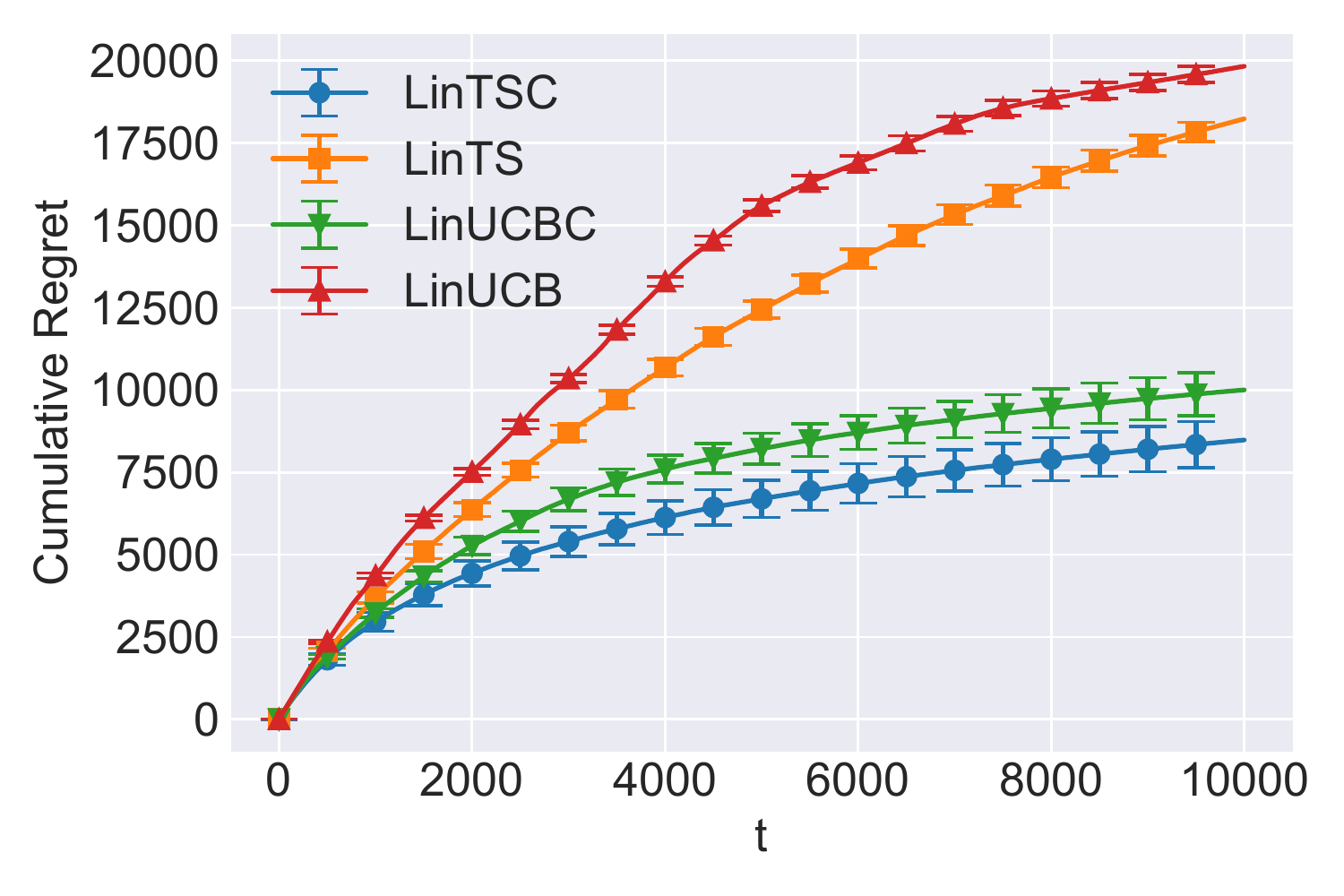}\label{fig:context_large_eps05}} &
\subfloat[CBC with $k=30$, $n=900$, $\epsilon=0.1$]{\includegraphics[width=0.3\textwidth]{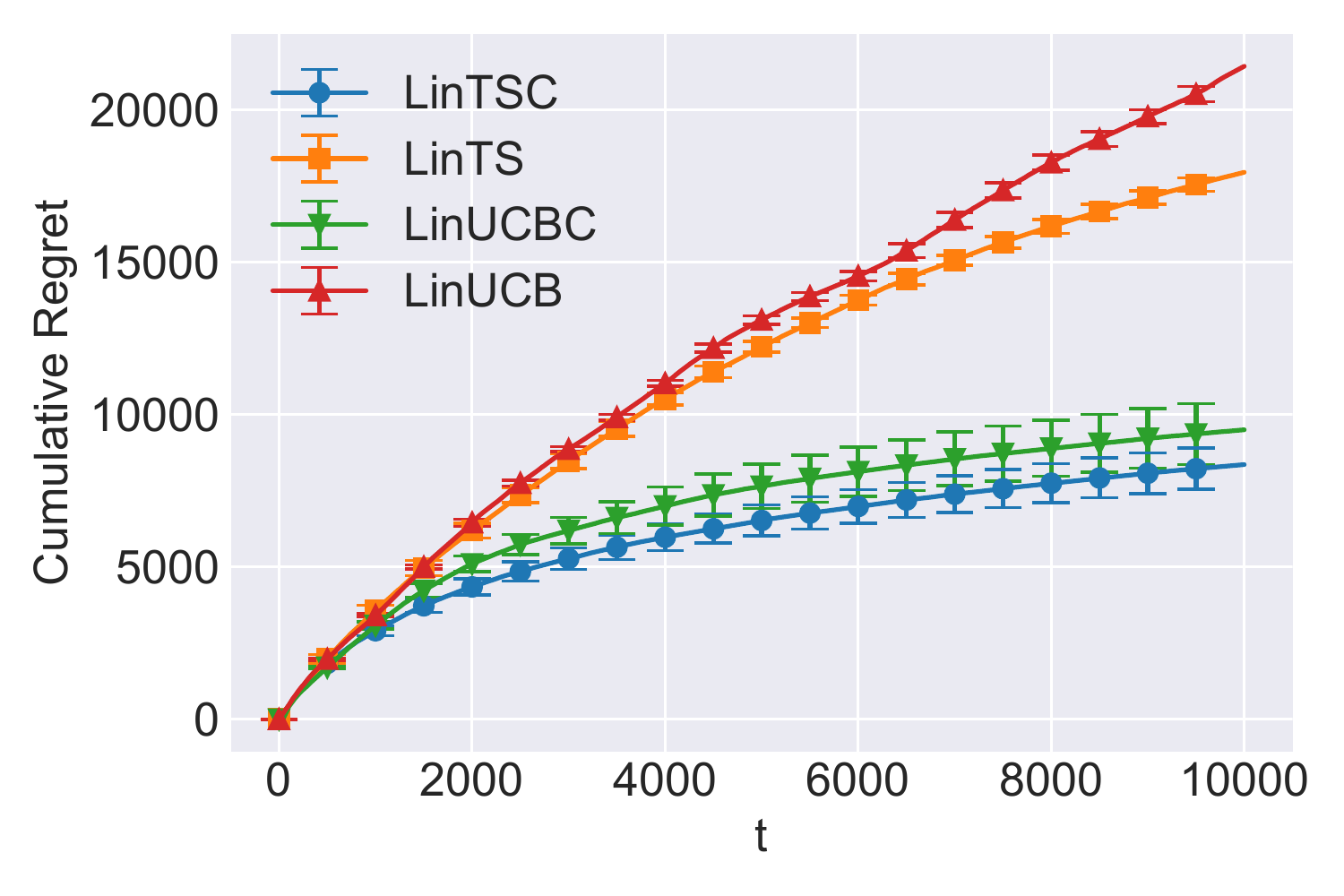}\label{fig:context_large_eps01}} 
\\
\end{tabular}
\caption{CBC and violation of assumptions in MABC.}
\end{figure*}

\paragraph{Violation of Assumptions.} In a real world setting, assuming that strong dominance and especially hierarchical strong dominance holds completely is often too strong. We thus evaluate our algorithms on instances for which these assumptions are violated. We generate $N$ arms by for each arm $i$ we sample a value $x_i \sim \mathcal{U}(0, 1)$. We cluster the arms into $K$ clusters, based on the values $\{x_i\}$, using K-means. The reward distribution of each arm $i$ is a Bernoulli distribution with mean $f(x_i)$ where $f(x) = \frac{1}{2}(\sin{13x}\sin{27x} + 1)$.
This function is illustrated in the supplementary material, Appendix A, and has previously been used to evaluate bandit algorithms in \citet{Bubeck11}, the smoothness of the function ensures arms within the same cluster to have similar expected rewards, on the other hand the periodicity of $\sin $ yields many local optima and the optimal cluster won't strongly dominate the other clusters. On these instances, we benchmark TSC against two another algorithms proposed for MABC, UCBC \citep{Pandey2007, Bouneffouf} and TSMax \citep{Zhao19}. We also benchmark against UCB1 \citep{Auer2002} and TS which both considers the problem as a standard MAB, making no use of the clustering. We run the algorithms on two different instances, one with $N=100$ and $K=10$ and the other one with $N=1000$ and $K=32$. For each instance we run the algorithms on $100$ different random seeds and we present the results in Figure: \ref{fig:kmeans_small} and \ref{fig:kmeans_large}, the error bars corresponds to $\pm 1$ standard deviation. TSC outperforms the other algorithms on both instances and especially on the larger instance where there is a big gap between the regret of TSC and the regret of the other algorithms. In order to test HTS we generate an instance, as above, with $N=5000$ and $K=15$ and construct a tree by recursively breaking each cluster up into $15$ smaller clusters using k-means. In Figure \ref{fig:HTS_UCT} we show the performance of HTS for two different levels, $L=2, 3$, compared to TSC using the clusters at level $L=1$ in the tree and also compared to the UCT-algorithm \citep{Kocsis} using the same levels of the tree as HTS. We averaged over $100$ random seeds. The HTS performs well on this problem and is slightly better than TSC while both HTS and TSC outperforms UCT. We present more empirical results for MABC in the supplementary material.

\subsection{Contextual Bandit}
We generate contextual data in the same way as in \citet{Bouneffouf}. We have $K$ clusters and $N$ arms. Each arm $j$ is randomly assigned to a cluster $i$. For each cluster $i$ we sample a centroid  $\theta_{i}^{c} \sim \mathcal{N}(0, \mathbf{1}_5)$ and define a coefficient for each arm $j$ in the cluster as $\theta_j = \theta_{i}^{c} + \epsilon v, \, v \sim  \mathcal{N}(0, \mathbf{1}_5)$. We take the reward of an arm to be $\mathcal{U}(0, 2 \theta_{j}^\intercal x)$ where $x$ is the given context. The reward becomes linear and we can control the expected diameter of a cluster by varying $\epsilon$.

We benchmark LinTSC against the UCB-based counterpart LinUCBC \citep{Bouneffouf} and the standard algorithms LinTS \citep{Agrawal12b} and LinUCB \citep{Li_2010}, which treats the problem as a standard CB. We ran the algorithms on three different instances presented in Figures \ref{fig:context_small}, \ref{fig:context_large_eps05} and \ref{fig:context_large_eps01}, over $25$ different random seeds and the error bars corresponds to $\pm 1$ standard deviation. We run all algorithms with there corresponding standard parameter ($v=1$ for LinTS and LinTSC, $c=2$ for LinUCB and LinUCBC).
We see a clear improvement between not using the clustering (TS) and using the clustering (TSC). LinTSC performs slightly better than LinUCBC as the problem becomes larger w.r.t number of arms and clusters, Figures \ref{fig:context_large_eps05} and \ref{fig:context_large_eps01}.

\section{Related Work}\label{sec:r_work}

Bandits are now a classical subject in machine learning and recent textbook treatments are \citet{Bubeck12, Slivkins2019, lattimore20}. The MABC and CBC can be considered as natural special cases of the more general finite-armed structured bandit which is studied in  \citep{Lattimore14, Combes17, Gupta2018, gupta2020multiarmed}. To the best of our knowledge, the idea of clustered arms was first studied in \citet{Pandey2007} and the MABC corresponds to their undiscounted MDP setup for which the authors propose a general two-level bandit policy and gives theoretical justifications on how the regret scales depending on the characteristics of the clustering, but without stating  rigorous regret bounds. Bandits with clustered arms were also recently studied in \citet{Bouneffouf, Matthieu2019} and both papers prove regret bounds for UCB-styled algorithms in the MABC under various assumptions on the clustering. \citet{Bouneffouf} is the work most related to ours since they consider a two-level UCB scheme and  regret bounds that exhibits similar dependence on the clustering quality as our bounds. In \citet{Zhao19} the authors propose a two-level  TS algorithm where the belief of a cluster is set to the belief of the best performing arm in the cluster so far and the authors give no theoretical analysis of its regret. Clustered arms also appear in the regional bandit model \citep{Wang18b, Singh20} under the assumption that all arms in one cluster share the same underlying parameter.  Another model related to our work is the latent bandit \citep{Maillard2014, Hong2020} where the reward distributions depends on a latent state and the goal of the learner is to identify this state. 

Bandits and tree structures are studied using a UCB-styled algorithm for Monte-Carlo-based planning in the influential work \citet{Kocsis} and later studied for various bandit problems with smoothness in the seminal works \citet{Coquelin07, Bubeck11}. 

We have based our bandit algorithms on the classical method Thompson sampling \citep{Thompson1933} which has been shown to perform well in practise \citep{Chapelle2011} and for which rigorous regret analyses recently have been established for the standard MAB in \citet{Kaufmann12, Agrawal2017}. The contextual version of Thompson sampling we use in our two-level scheme for CBC was originally proposed and analyzed for standard CB in \citet{Agrawal12b} and recently revisited in \citet{Abeille17}.

\section{Conclusions}\label{sec:conclusions}
In this paper, we have addressed the stochastic multi-armed bandit problem and the contextual bandit with clustered arms and proposed algorithms based on multi-level Thompson sampling.
We have shown that our algorithms can be used to drastically reduce the regret when a clustering of the arms is known and that these algorithms are competitive to its UCB-based counterparts. We think that the simplicity of our algorithms and the fact that one can easily incorporate prior knowledge makes them well-suited options for bandit problems with a known clustering structure of the arms. In the future we would like to explore how the regret of TSC behaves under weaker assumptions on the clustering. We want to determine what are sufficient properties of the clustering to ensure sub-linear regret of LinTSC. 
\section*{Acknowledgments}
We thank Tianchi Zhao for providing valuable comments on an earlier version of this paper. 

This work was supported by funding from CHAIR (Chalmers AI Research Center) and from the Wallenberg AI, Autonomous Systems and Software Program (WASP) funded by the Knut and Alice Wallenberg Foundation.  The computations in this work were enabled by resources provided by the Swedish National Infrastructure for Computing (SNIC).

\clearpage
\bibliography{refs}
\appendix
\label{appendix}
\section{Proofs}\label{app:proofs}
\subsection{Lemma \ref{lm:subopt_clusters}}
Assume $c^*=1$ is the cluster containing the optimal arm. We want to bound 
   $ \E[N_c] $
for some sub-optimal cluster $c$.  Let $\underline{\mu}_c$ be the smallest mean in cluster $c$ and let $\theta_{c, t}$ be the sample drawn from the belief of TSC for $c$ at time $t$.

If cluster $c$ is played at time $t$, i.e. $C_t=c$, then one of the two events need to happen \begin{itemize}
\item The sample, $\theta_{1, t}$ for cluster $1$ satisfy $$ \theta_{1, t} \leq \underline{\mu}_1 - \sqrt{\frac{6 \log t}{N_1}} $$
\item Or $ \theta_{1, t} > \underline{\mu}_1 - \sqrt{\frac{6 \log t}{N_1}} $ but $C_t=c$ anyway.
\end{itemize}
Thus the expected number of pulls, $N_c$, of  cluster $c$ can be decomposed as \begin{align}
    \E[N_{c, T}] \leq & \sum_{t=1}^T P\left(\theta_{1, t} \leq \underline{\mu}_1 - \sqrt{\frac{6 \log t}{N_{1, t}}}\right) \\
    & + \sum_{t=1}^T P\left(\theta_{c, t} > \underline{\mu}_1 - \sqrt{\frac{6 \log t}{N_{c, t}}}\right).
\end{align}

Let $\preccurlyeq$ denote stochastic domination, i.e. $X \preccurlyeq Y$ iff $P(X \geq x) \leq P(Y \geq x) \, \forall x$.  Let  $S_{c, t}$ and $F_{c, t}$ be the corresponding number of success and fail observations from cluster $c$ at time $t$, as in Algorithm \ref{alg:TSC}.  That is, \begin{align}
S_{c, t} &= 1 + \sum_i^{N_{c, t}} r_{i} \\
F_{c, t} &= 1+  \sum_i^{N_{c, t}} 1 - r_{i}
\end{align}
where $r_{i}$ is a reward drawn from some arm in cluster $c$.  

To bound the first term in the inequality,  $\sum_{t=1}^T P\left(\theta_{1, t} \leq \underline{\mu}_1 - \sqrt{\frac{6 \log t}{N_1}}\right) $,  we consider an auxiliary sample $\theta_{1, t}' \sim \text{Beta}(S_{1, t}', F_{1, t}')$ such that \begin{align}
S_{1, t}' &= 1 + \sum_i^{N_{1, t}} r_{i}' \\
F_{1, t}' &= 1+  \sum_i^{N_{1, t}} 1 - r_{i}'
\end{align}
where $r_i'$ corresponds to sample drawn from the worse arm in cluster $1$ with mean $ \underline{\mu}_1 $.  It is easy to verify that $S_{c, t}'  \preccurlyeq S_{c, t}$ and $F_{c, t}  \preccurlyeq F_{c, t}'$ and thus $\theta_{1, t}' \preccurlyeq \theta_{1, t}$ \footnote{To see this, we use the beta-binomial trick $F_{a,b}^\text{Beta} = 1 - F_{a+b-1, x}^{\text{Binomial}}(a-1)$ and note that if $a+b=c+d=q$ and $a\geq c$ then $F_{q-1, x}^{\text{Binomial}}(c-1) \leq F_{q-1, x}^{\text{Binomial}}(a-1)$ which gives, using the trick, $F_{a,b}^\text{Beta}(x) \leq F_{c,d}^\text{Beta}(x), \, \forall x \in [0, 1]$, which implies stochastic dominance.}.  Thus we have \begin{align*}
    P\left(\theta_{1, t} \leq \underline{\mu}_1 - \sqrt{\frac{6 \log t}{N_{1, t}}}\right) \leq 
    P\left(\theta_{1, t}' \leq \underline{\mu}_1 - \sqrt{\frac{6 \log t}{N_{1, t}}}\right)
\end{align*}
and using Lemma 1 from \citet{Kaufmann12} we can conclude that \begin{align}
   \sum_{t=1}^\infty  P\left(\theta_{1, t}' \leq \underline{\mu}_1 - \sqrt{\frac{6 \log t}{N_{1, t}}}\right) \leq Q < \infty
\end{align}
where $Q$ is some constant.

We proceed in similar fashion to bound \begin{align}
 \sum_{t=1}^T P\left(\theta_{c, t} > \underline{\mu}_1 - \sqrt{\frac{6 \log t}{N_c}}\right).
\end{align}
We note that \begin{align}
P\left(\theta_{c, t} > \underline{\mu}_1 - \sqrt{\frac{6 \log t}{N_{c, t}}}\right) \leq P\left(\theta_{c, t}'' > \underline{\mu}_1 - \sqrt{\frac{6 \log t}{N_{c, t}}}\right)
\end{align}
where $\theta_{c, t}'' \sim \text{Beta}(S_{c, t}'', F_{c, t}'')$ with \begin{align}
S_{c, t}'' &= 1 + \sum_i^{N_{c, t}} r_{i}'' \\
F_{c, t}'' &= 1+  \sum_i^{N_{c, t}} 1 - r_{i}''
\end{align}
where $r_i''$ are observations from the best arm in cluster $c$ with mean $\overline{\mu}_c$.  By the same reasoning as previously we get $\theta_{c, t} \preccurlyeq \theta_{c, t}'' $. Applying Theorem \ref{th:kaufmann} yields \begin{align}
\E[N_{c, T}] \leq(1 + \epsilon) \frac{\log T + \log \log T}{\KL(\overline{\mu}_c, \underline{\mu}_{c^*})} + O(1)
\end{align}
for $\epsilon > 0$.

\subsection{Theorem \ref{th:instance_dependent}}
We can decompose the regret into \begin{align*}
    E[R_T] = \sum_{C\neq C^*}\sum_{a \in C} \Delta_a \E[N_{a, T}] + \sum_{a \in C^*} \Delta_a \E[N_{a, T}]
\end{align*}
where the first term consider the regret suffered from playing sub-optimal clusters and the second term regret suffered from playing sub-optimal arms within the optimal cluster. The second term can be bounded by just applying Theorem \ref{th:kaufmann} for $\epsilon > 0$ \begin{align*}
    \sum_{a \in C^*} \Delta_a \E[N_{a, T}] \leq (1 + \epsilon) \sum_{a \in C^*} \frac{1}{\Delta_a} \log T + o(\log T).
\end{align*}

To bound the first term, consider sub-optimal cluster $C$ and let $N_{C, T}$ denote the number of times we play $C$. Let $a_C^*$ be the action with highest expected reward in $C$. Then for any other $a \in C, \, a\neq a_C^*$ we can bound the number of plays, $N_{a, T_{C, T}}$, by Theorem \ref{th:kaufmann} \begin{align*}
     &\E[N_{a_C, N_{C, T}}] \leq (1 + \epsilon) \frac{1}{\KL(\mu_a, \mu_{a^*})}(\log N_{C, T} + \log \log N_{C, T}) \\ 
     & + O(1)
\end{align*}
and for $a_C^*$ we have 
\begin{align*}
    \E[N_{a_C^*, N_{C, T}}] \leq \E[N_{C, T}].
\end{align*}
From Lemma \ref{lm:subopt_clusters} we know that for $\epsilon > 0$ \begin{align*}
    \E[N_{C, T}] \leq (1 + \epsilon)\frac{1}{\KL(\overline{\mu}_C, \underline{\mu}_{C^*})}(\log T + \log \log T) + O(1)
\end{align*}

and we thus get a $\log \log T$ dependence on all arms in $C$ except the one with highest expected reward \begin{align*}
    & \E[N_{a, N_{C, T}}] \leq (1 + \epsilon)\frac{1}{\KL(\mu_a, \mu_{a^*})} \log \log T + o(\log \log T) \\
    &  \E[N_{a_C^*, N_{C, T}}] \leq (1 + \epsilon)\frac{1}{\KL(\overline{\mu}_C, \underline{\mu}_{C^*})}\log T + o(\log T).
\end{align*}

Therefore we can bound the regret suffered from sub-optimal clusters for any $\epsilon > 0 $ as \begin{align*}
    &\sum_{C\neq C^*}\sum_{a \in C} \Delta_a \E[N_{a, T}]\\
    &\leq (1 + \epsilon)(\sum_{C\neq C^*} \frac{\Delta_C }{\KL(\overline{\mu}_C, \underline{\mu}_{C^*})}\log T  + \\
    & + \sum_{a \in C, a\neq a^*} \frac{\Delta_a}{\KL(\mu_a, \mu_{a^*})}\log \log T)  + o(\log T) \\
    & \leq (1 + \epsilon)\sum_{C\neq C^*} \frac{\Delta_C }{\KL(\overline{\mu}_C, \underline{\mu}_{C^*})}\log T + o(\log T). 
\end{align*}

Combining with the bound on regret within the optimal cluster $C^*$ yields the instance-dependent regret bound \begin{align*}
    & \E[R_T] \leq \\
    & \leq (1+\epsilon)\left(\sum_{C\neq C^*} \frac{\Delta_C}{\KL(\overline{\mu}_C, \underline{\mu}_{C^*})}  +  \sum_{a \in C^*} \frac{\Delta_a}{\KL(\mu_a, \mu^*)}\right) \log T \\
    &+ o(\log T).
\end{align*}

\subsection{Theorem \ref{th:instance_independent}}
We rewrite $\Delta_C = d_C + w^*$  where $w^*$ is the width of the optimal cluster and hence by the definition of $\gamma_C$ we have \begin{align*}
    \Delta_C = (1 + \gamma_C) d_C.
\end{align*}
By Pinsker's inequality we have \begin{align*}
    \KL(\overline{\mu}_C, \underline{\mu}_{C^*}) \geq 2 d_C^2
\end{align*}
and for arms in the optimal cluster we have \begin{align*}
    \KL(\mu_a, \mu^*) \geq 2 \Delta_a^2
\end{align*}

Thus, the instance-dependent regret bound can be upper-bounded by \begin{align*}
 \frac{1+\epsilon}{2}\left(\sum_{C\neq C^*} \frac{1 + \gamma_C}{d_C}  +  \sum_{a \in C^*} \frac{1}{ \Delta_a}\right) \log T + o(\log T). 
\end{align*}

Let $\Delta > 0$. 
\begin{itemize}
    \item For all clusters $C$ and arms $a \in C^*$ such that $d_C, \Delta_a < \Delta$, the cumulative regret from these are upper-bounded by $\Delta T$. 
    \item For each cluster $C$ such that $d_C \geq \Delta$ the amount of regret suffered from playing $C$ is $O(\frac{1 + \gamma_C}{\Delta}\log T)$ and for each $a \in C^*$ the regret suffered is $O(\frac{1}{\Delta} \log T)$. In total this is $O(\frac{A^* + K(1 + \gamma)}{\Delta} \log T)$. 
\end{itemize}

Combining this yields \begin{align*}
    \E[R_T] \leq O(\Delta T + \frac{A^* + K(1 + \gamma) }{\Delta} \log T).
\end{align*}
Since this holds $\forall \Delta > 0$ we pick $\Delta = \sqrt{\frac{(A + K(1 + \gamma))\log T}{T}}$ and hence, \begin{align*}
    \E[R_T] \leq O\left(\sqrt{(A^* + K(1 + \gamma))T \log T}\right).
\end{align*}

\subsection{Theorem \ref{th:lower_bound}}
 We make use of the pioneering work of \cite{Lai1985} which gives that \begin{align}\label{eq:Lai}
     \lim_{T \xrightarrow{} \infty} \inf \frac{\E[R_T]}{\log T} \geq \sum_{a} \frac{\Delta_a}{\KL(\mu_a, \mu^*)}
 \end{align}
 for a standard multi-armed bandit with Bernoulli rewards. We can decompose the regret over sub-optimal clusters and sub-optimal arms in the optimal cluster \begin{align*}
     \E[R_T] = \sum_{C\neq C^*} \sum_{a \in C} \Delta_a \E[N_{a,T}] + \sum_{a \in C^*}\Delta_a \E[N_{a,T}], 
 \end{align*}
 and using the fact that the regret suffered within a sub-optimal cluster is bounded from below by the smallest regret in the cluster \begin{align*}
    \sum_{a \in C} \Delta_a \E[N_{a,T}]  \geq \Delta_C \sum_{a \in C} \E[N_{a, T}].
 \end{align*}
 Now we get the proposed bound by independently bounding each term from below by Equation:\ref{eq:Lai} and using the fact that for any cluster $C$ and any arm $a \in C$ we have \begin{align*}
     \KL(\mu_a, \mu^*) \geq \KL(\underline{\mu}_C, \mu^*).
 \end{align*}
 
 \subsection{Theorem \ref{th:independent_lower}}
  First consider the case where all arms are assigned to the same cluster. Any algorithm needs to at least have a $\sqrt{A^*T}$ dependence in the regret otherwise the lower bound $\Omega(\sqrt{NT})$ would be violated. 
 
 Secondly, consider the case where all clusters only contain one arm each. We have that any algorithm needs at least a $\sqrt{KT}$ dependence otherwise  $\Omega(\sqrt{NT})$ would be violated. 
 
 Since $\sqrt{K + A^*} \leq \sqrt{K} + \sqrt{A^*}$ it follows that for any algorithm we have \begin{align*}
     \E[R_T] \geq \Omega(\sqrt{(A^* + K)T}).
 \end{align*}
 
 \subsection{Theorem \ref{thm:HTS}}
   We decompose the cumulative regret into \begin{align*}
     R_T := \sum_{\mathcal{T}_{1}^j \neq \mathcal{T}_{1}^*}\sum_{a \in \mathcal{T}_{1}^j} \Delta_a \E[N_{a,T}] + \sum_{a \in \mathcal{T}_{1}^*} \Delta_a \E[N_{a,T}].
 \end{align*}
 Since strong dominance holds on each level we bound the first sum by $\sum_{j=2}^{K_{0}^{*}}\frac{\Delta_{1}^j}{(2d_{1}^j)^2}\log T + o(\log T)$ using Theorem \ref{th:instance_dependent}, where $(2d_{1}^j)^2$ follows from Pinsker's inequality for Bernoulli distributions. We are left with bounding the regret from \begin{align*}
     \sum_{a \in \mathcal{T}_{1}^*} \Delta_a \E[N_{a,T}] = \sum_{j=2}^{K_1^*}\sum_{a \in \mathcal{T}_{2}^j} \Delta_a \E[N_{a,T}] + \sum_{a \in \mathcal{T}_{2}^*} \Delta_a \E[N_{a,T}].
 \end{align*}
 And we recursively apply Theorem \ref{th:instance_dependent} to bound the first time like above, until we reach level $L$ for which we use Theorem \ref{th:kaufmann} along with Pinsker's inequality to get \begin{align*}
    \sum_{a \in \mathcal{T}_{L}^*} \Delta_a \E[N_{a,T}] \leq (1 + \epsilon) \sum \frac{1}{\Delta_a} \log T + o(\log T)
 \end{align*}
 
\section{Empirical Evaluation MABC}\label{app:MABC}
\begin{figure}[h]
\centering
\begin{tabular}{c}
\subfloat[The function $f(x)=\frac{1}{2}(\sin{13x}\sin{27x} + 1)$ used for expected rewards in the evaluation of TSC and HTS with violated assumptions in Section \ref{sec:results}.]{\includegraphics[scale=0.5]{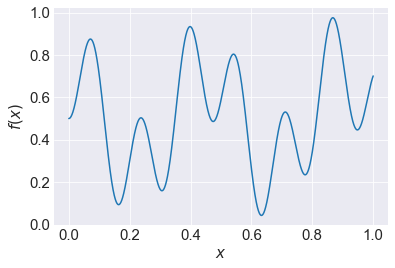}}\\
\subfloat[$f(x) =\frac{1}{2}(e^{-\frac{1}{0.05}(0.1-x)^2}+e^{-\frac{1}{0.8}(0.9-x)^2})$ used for expected rewards in Figure \ref{fig:gaussian_MABC}.]{\includegraphics[scale=0.5]{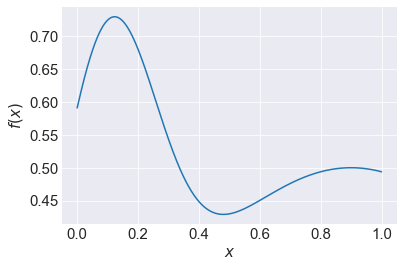}\label{fig:gaussian}} \\
\subfloat[2 dimensional function used for expected rewards in Figure \ref{fig:2d-MABC}.]{\includegraphics[scale=0.5]{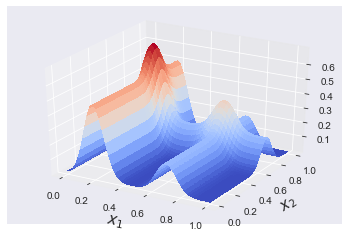}\label{fig:2D_func}}
\end{tabular}
\caption{Functions used for evaluating TSC and HTS when theoretical assumptions are violated.}
\end{figure}

\begin{figure}[h]
\centering
\begin{tabular}{c}
\subfloat[Cumulative regret over the 2-d instance with $N=500$ and $K=20$.]{\includegraphics[scale=0.5]{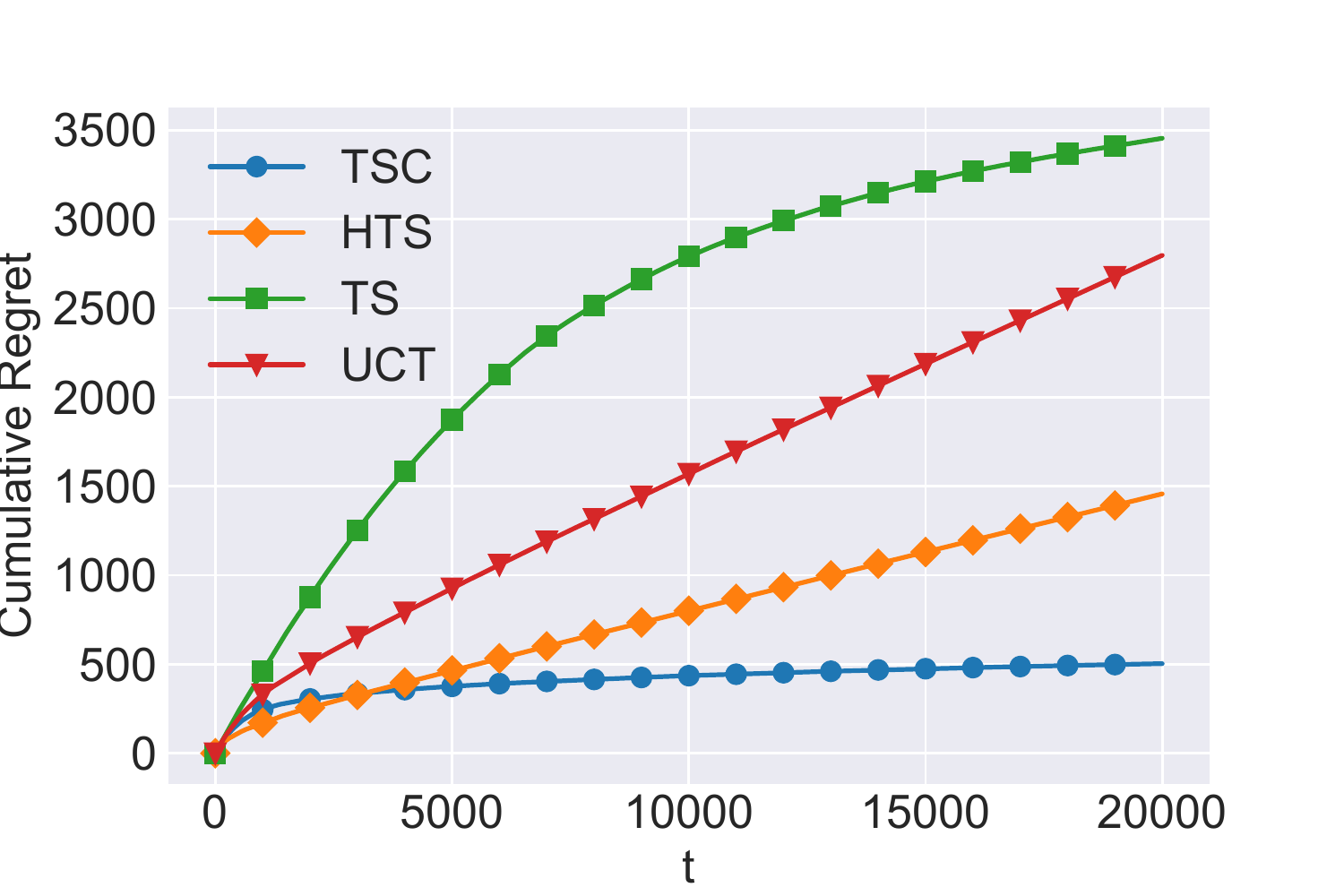}\label{fig:2d-MABC}}\\
\subfloat[Cumulative regret over the instance with $N=50$ arms with expected rewards as $f(x)$ in Figure \ref{fig:gaussian}.]{\includegraphics[scale=0.5]{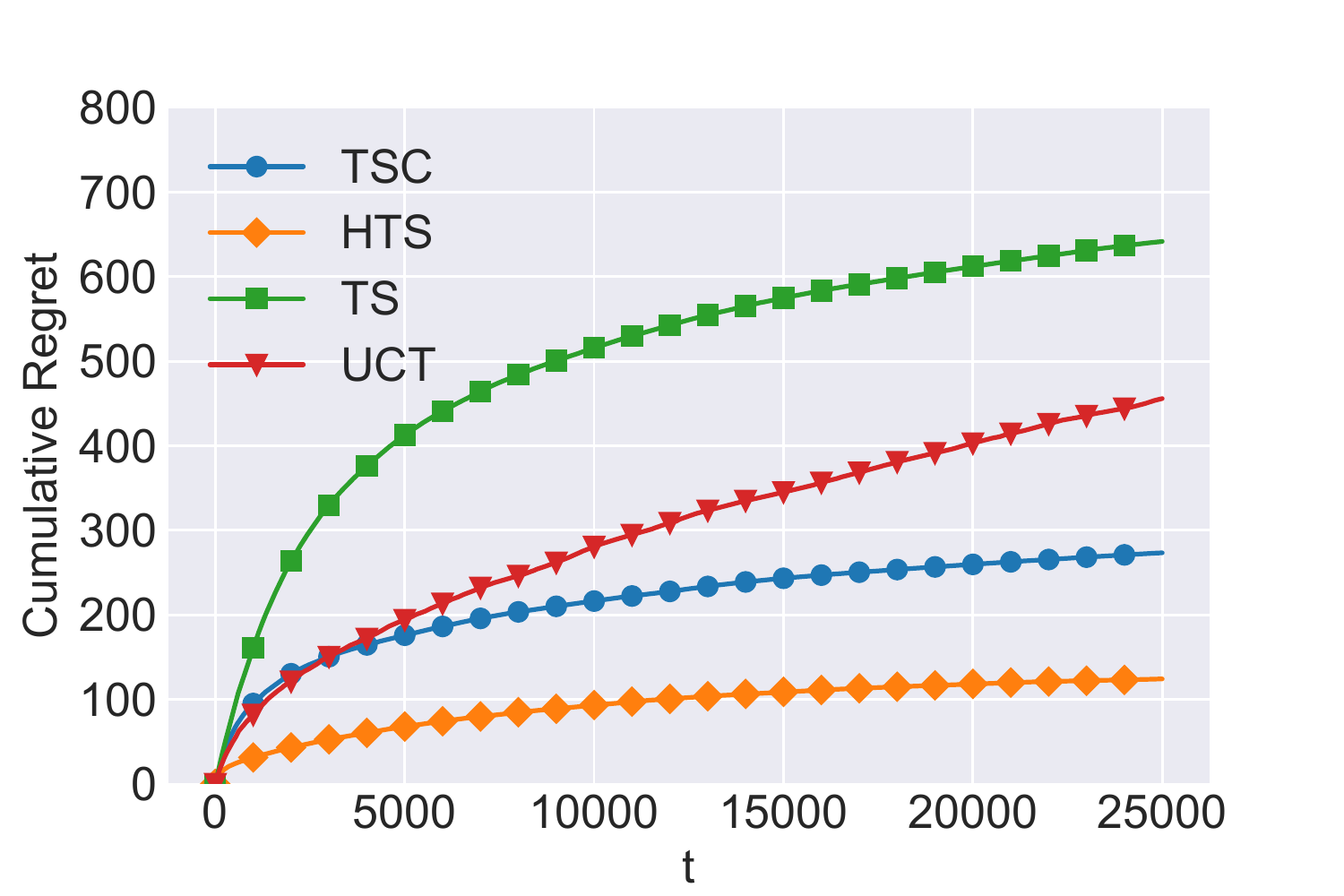}\label{fig:gaussian_MABC}} \\
\subfloat[Cumulative regret over uniformly assigned expected rewards. $N=50$ and $K=10$.]{\includegraphics[scale=0.5]{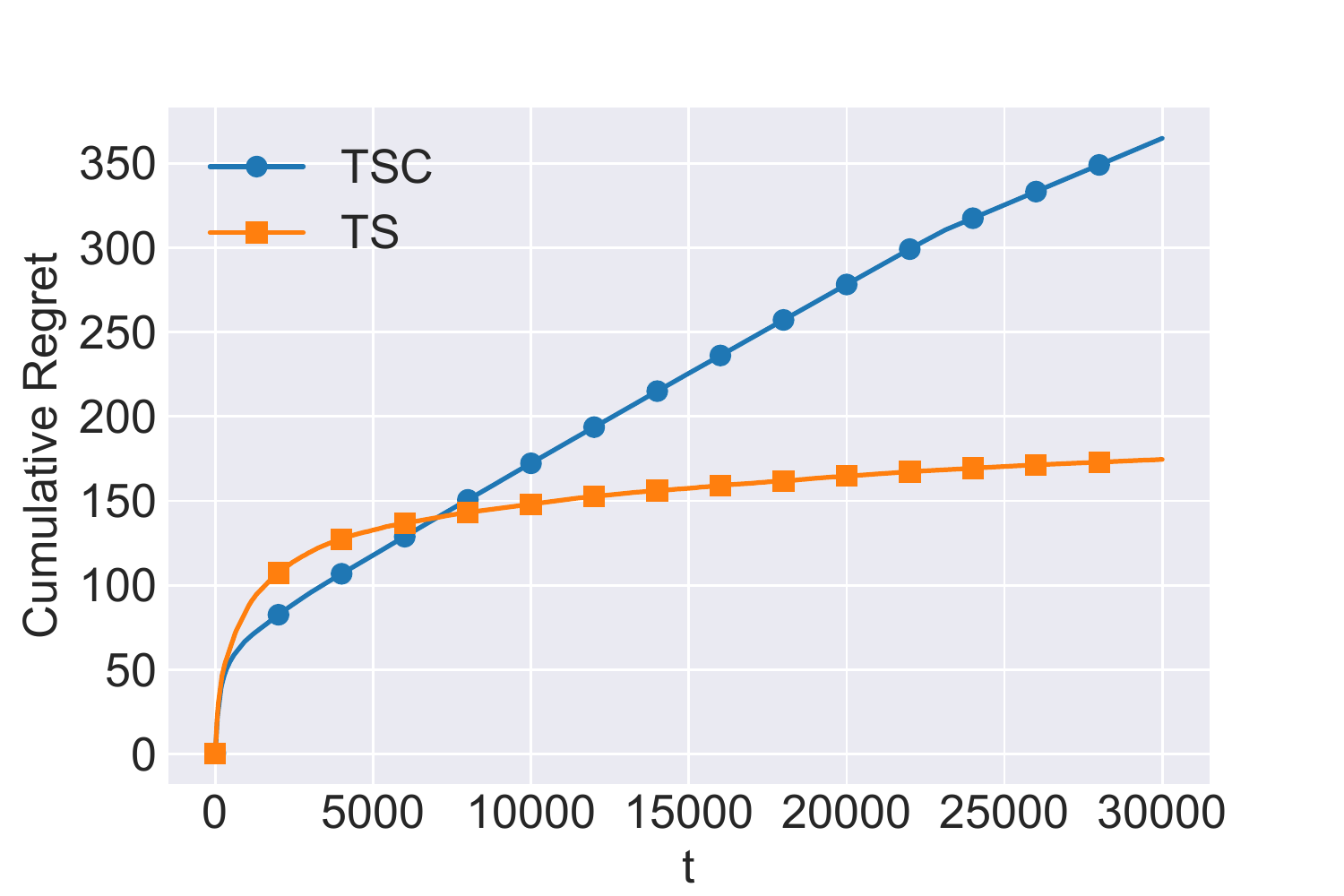}\label{fig:random_clustering}}
\end{tabular}
\caption{}
\end{figure}

To give an example where HTS achieves linear regret while TSC exhibits sub-linear regret we have $N=500$ arms and for each arm $a_i$ we draw a vector $x_i$ from $x_i \sim \mathcal{U}([0, 1]^2)$. We cluster the arms into $K=20$ clusters using k-means and use that clustering in TSC. We also cluster the arms using agglomerative clustering and use the resulting tree for HTS and UCT. We take the reward for each arm $a_i$ to be Bernoulli distributed with mean reward
\begin{align*}
    f(x_1, x_2) = & \frac{1}{2}e^{-100(0.2-x_1)^2} + \frac{1}{5}e^{-100(0.7-x_1)^2} + \\
    & \frac{1}{5}e^{-100(0.7-x_2)^2},
\end{align*}
this function is illustrated in Figure \ref{fig:2D_func}. This function is chosen such that there is a similarity between close arms but as we go higher up in the tree arms in the same sub-tree may have very different rewards. We run the algorithms for $T=20 \, 000$ and over $25$ random seeds and in Figure \ref{fig:2d-MABC} we see that both UCT and HTS exhibits linear cumulative regret curve while TSC is still sub-linear since arms clustered together tends to have similar reward. Hence, using the full tree in this case is a too aggressive exploration scheme and we see that care has to be taken in HTS when deciding how deep the hierarchical clustering should be. 

We also generated a bandit instance using the function \begin{align*}
    f(x) = \frac{1}{2}(e^{-\frac{1}{0.05}(0.1-x)^2}+e^{-\frac{1}{0.8}(0.9-x)^2}),
\end{align*}
illustrated in Figure \ref{fig:gaussian}. This function is considered since it is very smooth and one may assume similar rewards for arms in the same sub-tree of a hierarchical clustering. We generate $N=50$ arms as before and for TSC we cluster them using k-means with $K=5$. For HTS and UCT we use agglomerative clustering and consider the full tree. We run the algorithms for $T=25 \, 000$ and over $25$ random seeds and present the results in Figure \ref{fig:gaussian_MABC}. We see that for this instance HTS exhibits sub-linear regret and performs better than TSC, for this clustering. This illustrate that the quality of the clustering is very important for the regret, especially for HTS. 

We also compare TS and TSC on an instance where there is no correlation between rewards in a cluster. We take $N=50$ arms and divide them into $K=10$ clusters. The reward of each arm is Bernoulli distributed and we draw the expected reward of each arm from $\mathcal{U}(0, 1)$. The average over $25$ random seeds is presented in Figure \ref{fig:random_clustering} and as expected we see that TSC has a cumulative regret which is worse then TS, since the quality of the clustering is bad. 

\end{document}